\documentclass[sigconf]{acmart}

\usepackage[utf8]{inputenc} % allow utf-8 input
\usepackage[T1]{fontenc}    % use 8-bit T1 fonts
\usepackage{xcolor}
\definecolor{mydarkblue}{rgb}{0,0.08,0.45}
\definecolor{mydarkgreen}{rgb}{0,0.45,0.15}

\usepackage{url}            % simple URL typesetting
\usepackage{booktabs}       % professional-quality tables
\usepackage{amsfonts}       % blackboard math symbols
\usepackage{nicefrac}       % compact symbols for 1/2, etc.
\usepackage{microtype}      % microtypography
\usepackage{graphicx}
\usepackage{subcaption}     % subcaptions
\usepackage{array}
\usepackage{fancyvrb}
\usepackage{alltt}
\usepackage{fancybox}
\usepackage{pifont}
\usepackage{verbatim} 
\usepackage{hyperref}
\definecolor{Gray}{gray}{0.97}
\definecolor{lightgray}{gray}{0.9}
\definecolor{Purple}{rgb}{0.7,0.0,0.7}
\definecolor{Orange}{rgb}{1, 0.5, 0.1}
\definecolor{OliveGreen}{rgb}{0.36, 0.71, 0.39}
\definecolor{Blue}{RGB}{3, 94, 255}
\definecolor{Red}{RGB}{255, 1, 0}

\AtBeginDocument{%
  }

\copyrightyear{2025}
\acmYear{2025}
\setcopyright{acmlicensed}
\acmConference[CHI '25]{CHI Conference on Human Factors in Computing Systems}{April 26-May 1, 2025}{Yokohama, Japan}
\acmBooktitle{CHI Conference on Human Factors in Computing Systems (CHI '25), April 26-May 1, 2025, Yokohama, Japan}
\acmDOI{10.1145/3706598. 3714096}
\acmISBN{979-8-4007-1394-1/25/ 04}

\begin{document}

\title{VideoA11y: Method and Dataset for Accessible Video Description}

\author{Chaoyu Li}
\affiliation{%
  \department{School of Computing and Augmented Intelligence}
  \institution{Arizona State University}
  \city{Tempe}
  \state{Arizona}
  \country{USA}
}
\email{chaoyuli@asu.edu}
\author{Sid Padmanabhuni}
\affiliation{%
  \department{School of Computing and Augmented Intelligence}
  \institution{Arizona State University}
  \city{Tempe}
  \state{Arizona}
  \country{USA}
}
\email{spadma20@asu.edu}
\author{Maryam Cheema}
\affiliation{%
  \department{School of Computing and Augmented Intelligence}
  \institution{Arizona State University}
  \city{Tempe}
  \state{Arizona}
  \country{USA}
}
\email{mcheema2@asu.edu}
\author{Hasti Seifi}
\affiliation{%
  \department{School of Computing and Augmented Intelligence}
  \institution{Arizona State University}
  \city{Tempe}
  \state{Arizona}
  \country{USA}
}
\email{hasti.seifi@asu.edu}
\author{Pooyan Fazli}
\affiliation{%
  \department{School of Arts, Media and
Engineering}
  \institution{Arizona State University}
  \city{Tempe}
  \state{Arizona}
  \country{USA}
}
\email{pooyan@asu.edu}


\begin{abstract}%150 words
  Video descriptions are crucial for blind and low vision (BLV) users to access visual content. However, current artificial intelligence models for generating descriptions often fall short due to limitations in the quality of human annotations within training datasets, resulting in descriptions that do not fully meet BLV users' needs. To address this gap, we introduce VideoA11y, an approach that leverages multimodal large language models (MLLMs) and video accessibility guidelines to generate descriptions tailored for BLV individuals. Using this method, we have curated VideoA11y-40K, the largest and most comprehensive dataset of 40,000 videos described for BLV users. Rigorous experiments across 15 video categories, involving 347 sighted participants, 40 BLV participants, and seven professional describers, showed that VideoA11y descriptions outperform novice human annotations and are comparable to trained human annotations in clarity, accuracy, objectivity, descriptiveness, and user satisfaction. We evaluated models on VideoA11y-40K using both standard and custom metrics, demonstrating that MLLMs fine-tuned on this dataset produce high-quality accessible descriptions. Code and dataset are available at \url{https://people-robots.github.io/VideoA11y/}.
  
\end{abstract}




\maketitle

\section{Introduction} 
New video content is created at an astounding rate, further widening the digital accessibility (a11y) gap experienced by blind and low vision (BLV) people. Video description, also known as audio description (AD), can make videos accessible to BLV users by narrating the visual content of a scene, such as actions, characters, scene changes, and interactions~\citep{usaccess,hitl_blv_2020, bodi_2021,cheema2024nowuser}. For professionally created media, such as films and television shows, producing ADs requires significant collaborative efforts from a team of experts, including producers, audio description writers, voice actors, and audio engineers~\cite{fcc}. Thus, smaller studios and independent films may not always provide AD. For user-generated content, which has surged in popularity on platforms such as YouTube and TikTok, the implementation of ADs lags considerably behind~\cite{review_hci}. YouDescribe~\cite{youdescribe} is an online platform where users can record and upload descriptions for YouTube videos. However, most videos in its wish list remain undescribed since the time, training, and confidence needed to create quality descriptions can deter potential contributors~\citep{image_descriptions_2015, hitl_blv_2020}. Given the rapid increase in online videos, human description alone is insufficient, making artificial intelligence (AI)-generated audio descriptions a viable alternative.

\begin{figure*}
    \centering
    \includegraphics[width=0.85\linewidth]{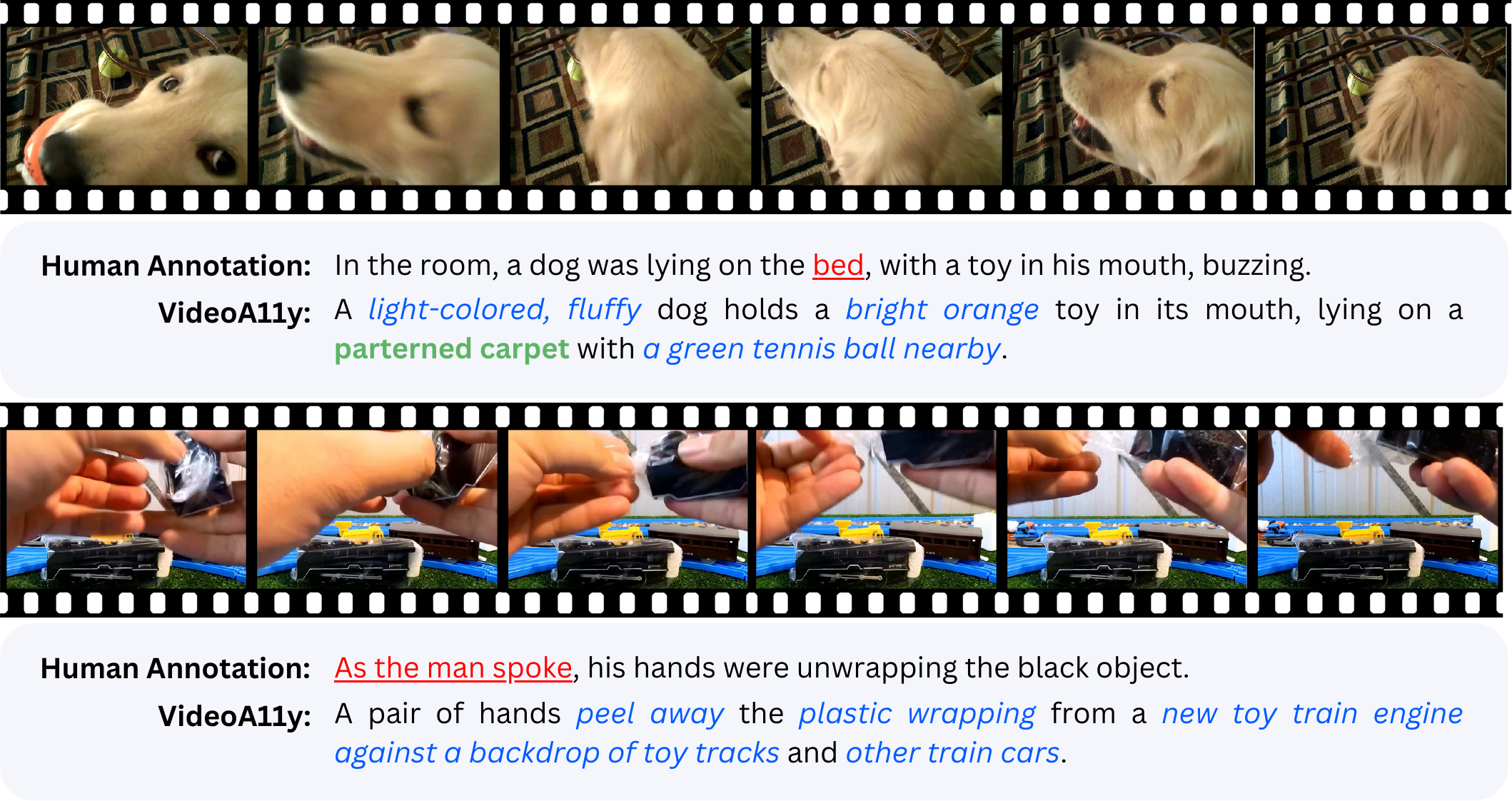}
    \caption{The human annotations and descriptions generated by VideoA11y for six consecutive frames of two sample video clips. \textcolor{Red}{\underline{Red underline}} indicates the errors in human annotations, \textcolor{OliveGreen}{\textbf{green bold}} indicates the corrected facts, and \textit{\textcolor{Blue}{blue italics}} indicates additional details.}
    \Description{Comparisons between human annotations and descriptions generated by VideoA11y for two sample video clips, each depicted through six frames as a sequence. The descriptions highlight differences where VideoA11y provides additional or corrected details. The top row showcases a sequence involving a dog with differences in descriptive accuracy between the two methods, while the bottom row features a scene involving hands unwrapping a toy, where VideoA11y again adds detail and accuracy to the description. For the first example, the human annotation is: "In the room, a dog was lying on the bed, with a toy in his mouth, buzzing.", where the phrase "bed" is underlined in red. VideoA11y's description is: "A light-colored, fluffy dog holds a bright orange toy in its mouth, lying on a patterned carpet with a green tennis ball nearby.", where the phrases "light-colored, fluffy", "bright orange", and "a green tennis ball nearby" are italics in blue, and the phrase "patterned carpet" is bold in green. For the second example, the human annotation is: "As the man spoke, his hands were unwrapping the black object.", where the phrase "As the man spoke" is underlined in red. VideoA11y's description is: "A pair of hands peel away the plastic wrapping from a new toy train engine against a backdrop of toy tracks and other train cars.", where the phrases "peel away", "new toy train engine against a backdrop of toy tracks", and "other train cars" are italics in blue.}
    \label{fig:examples}
\end{figure*}

In recent years, advances in computer vision and natural language processing (NLP) have enabled the development of new techniques for automatically generating video descriptions~\citep{chuang2023clearvid} using multimodal large language models (MLLMs)~\citep{videollava,vila}. These models are typically trained on general video description datasets, which include videos paired with descriptions or annotations (we use the terms `description' and `annotation' interchangeably) created by either humans or AI. However, existing datasets are insufficient for generating descriptions that effectively support BLV individuals in understanding video content. A key limitation is that annotations in these datasets often contain errors and fail to adhere to AD guidelines for accessibility. Human-generated descriptions also tend to be brief and can contain grammatical, spelling, or semantic errors, which may limit video comprehension for BLV audiences.

To address this gap, we introduce the VideoA11y method and the VideoA11y-40K dataset. VideoA11y is a novel approach designed to generate high-quality descriptions from scratch or enhance existing annotations for a wide range of video categories. It aims to produce detailed and accurate descriptions, thereby improving content accessibility for BLV users. To this end, we have compiled and summarized 42 guidelines from professional AD sources that capture the needs of BLV individuals. We then leveraged MLLMs to generate accessible descriptions using a prompt that adheres to these guidelines (i.e., compliant prompt). Figure \ref{fig:examples} shows examples of human annotations and revised descriptions generated by VideoA11y using GPT-4 Vision (GPT-4V)~\cite{openai2024gpt4} as the MLLM. We used VideoA11y to curate VideoA11y-40K, the largest and most comprehensive video description dataset for training accessible models. The dataset includes 40,000 videos across 15 categories, all specifically described for BLV individuals. To evaluate VideoA11y and VideoA11y-40K, we asked the following research questions:

\begin{enumerate}

    \item \textbf{RQ1.} How do VideoA11y descriptions compare in quality to those created by novice and trained human describers?
 
    \item \textbf{RQ2.} How do professional describers and BLV users evaluate and prefer VideoA11y descriptions compared to human descriptions?
    \item \textbf{RQ3.} Can the VideoA11y-40K dataset enhance state-of-the-art (SOTA) open-source MLLMs to generate high-quality video descriptions specifically tailored for BLV individuals? 
\end{enumerate}

To answer these questions, we conducted five user studies with both sighted and BLV individuals. Study 1 evaluated both open-source and proprietary MLLMs to determine the best model for VideoA11y. A group of 150 sighted users on Amazon Mechanical Turk (MTurk) watched 150 videos from 15 categories (e.g., education, sports) and rated descriptions generated by VideoA11y using these MLLMs on four evaluation metrics: descriptiveness, objectivity, accuracy, and clarity. After selecting the most suitable model, the subsequent four studies focused on assessing the effectiveness of the VideoA11y method and the VideoA11y-40K dataset. In Study 2, another 150 sighted MTurk users watched the same 150 videos from Study 1 and rated descriptions generated by VideoA11y or novice humans on the same four metrics. In Study 3, 47 sighted MTurk users watched 47 YouTube videos and rated descriptions generated by VideoA11y and high-quality annotations produced by four members of our team, following the 42 AD guidelines. In Study 4, seven professional audio describers evaluated the same 47 videos and descriptions, selecting the better description for each video to further assess the quality of the annotations. Study 5 evaluated the alignment between video descriptions and the needs and satisfaction of BLV users. Specifically, 40 BLV users watched 10 videos from five categories with human-generated and VideoA11y descriptions. They rated each description based on the four metrics, selected their preferred description, and provided reasons for their choice. The results of these studies demonstrate that VideoA11y produces video descriptions of superior quality in all metrics compared to novice human annotations and is comparable to the quality of annotations produced by trained humans. Finally, we developed two complementary benchmarks to evaluate open-source MLLMs on VideoA11y and VideoA11y-40K, using standard NLP metrics and the four custom metrics of descriptiveness, objectivity, accuracy, and clarity. Our work is pioneering in the HCI and AI community, focusing on creating a video description dataset specifically for BLV users and validating it with both sighted and BLV individuals. The novelty of this work lies in bridging established human practices of audio description with advancements in video description models and in creating a method, dataset, and benchmark dedicated to video accessibility. Our user studies and benchmark experiments demonstrate that MLLM-generated descriptions not only surpass the quality of novice human annotations but are also comparable to the standards of trained human annotations for video accessibility. In summary, the contributions of this paper are as follows: 
\begin{itemize}
    \item %\textbf{Prompt with Guidelines.} 
    Develop VideoA11y, an MLLM-based approach for generating video descriptions using 42 AD guidelines that we collated to focus on the needs of BLV individuals. 
    
   \item 
   Release the first and most comprehensive video description dataset, VideoA11y-40K, for training models for BLV users. 

    \item 
    Demonstrate the effectiveness of VideoA11y and VideoA11y-40K via evaluation studies with 347 sighted participants, 40 BLV participants, and 7 professional audio describers. 
    
    \item Introduce a new benchmark for video accessibility based on VideoA11y-40K. 
 
\end{itemize}

\section{Related Work}
We review prior work on video accessibility, MLLMs for video understanding, and video description datasets and metrics.

\subsection{Interactive Systems for Video Accessibility}

Prior work on video accessibility aimed to simplify the task of sighted describers in writing descriptions~\citep{lei-etal-2020-mart, kobayashi2010synthesized, video_scene_natalie_2022}. LiveDescribe~\cite{livedescribe} developed an interface for novice volunteers to create audio descriptions. Similarly, Rescribe~\cite{rescribe} assisted authors in creating and timing audio descriptions by optimizing content length and enabling iterative adjustments using dynamic programming. CrossA11y~\cite{crossa11y} further supports AD authors by detecting visual and auditory accessibility issues in videos, using cross-modal grounding analysis and an interface for reviewing and refining audio descriptions. However, these tools cannot generate partial or complete descriptions automatically. To address this issue, Yuksel et al.~\cite{hitl_blv_2020,yuksel2020increasing} developed a human-in-the-loop machine learning approach in which the AI system provides initial video descriptions, and sighted novices edit the descriptions to enhance their quality. This approach improved the quality of video descriptions while reducing the time and effort required from volunteer describers. Yet, it still requires manual editing, making it difficult to scale. In response, Bodi et al.~\cite{bodi_2021,ihorn2021narrationbot} developed a fully automated system that generates descriptions and enables interactive question answering based on the visual content of a video.

More recent work used LLMs to summarize AI-generated descriptions for individual keyframes in a video. For example, ShortScribe~\cite{shortscribe} leveraged automatic speech recognition (ASR), the image captioning model BLIP2~\cite{blip2}, and optical character recognition (OCR) to generate descriptions of several keyframes in a video, which are then summarized by GPT-4 to produce a video description. Similarly, SPICA~\cite{spica} uses an image captioning model~\cite{ofa} to describe keyframes, followed by GPT-4 to turn the descriptions into a coherent narrative. These methods follow a hierarchical structure, generating descriptions for static frames first and then merging the descriptions with an LLM. This approach can result in missed context, inaccuracies during temporal changes, and semantic inconsistencies in the descriptions. In contrast, VideoA11y leverages MLLMs to process keyframes and generate video descriptions, preserving temporal information and minimizing semantic loss.

\subsection{Multimodal Large Language Models for Video Understanding}

Recent MLLMs demonstrate outstanding abilities in understanding, interpreting, and analyzing video content. MLLMs are trained on large multimodal (e.g., video, audio, text) datasets~\citep{vast, howto100m, yt-temporal}, then are fine-tuned or instruction-tuned for specific tasks. Fine-tuning involves taking a pre-trained model and training it further on a smaller, task-specific dataset. In video understanding, this could involve using a model pre-trained on general multimodal datasets and adapting it to tasks like video description or video question answering~\citep{zhao2023lavila, vast, vid2seq, merlin}. Fine-tuning results in a highly specialized model for that task but may reduce its generalization capabilities across other tasks. On the other hand, instruction tuning enhances a model’s ability to generalize across various tasks by improving how well it follows instructions~\citep{videollama, videollava, vtimellm, gpt4video, VideoChatGPT}. The model is adjusted using diverse instructions that teach it how to interpret and perform different tasks. Our work leverages pre-trained MLLMs, which are subsequently fine-tuned on the proposed VideoA11y-40K dataset to benchmark their performance in generating accessible video descriptions for BLV users.

Other research has explored the use of prompt engineering to enhance model performance~\citep{gpt2, fewshot, cot_prompt, optimization_prompt}. Prompt engineering involves designing and optimizing inputs (prompts) to guide the model in generating relevant and accurate outputs without additional training on the pre-trained model. Previous work~\cite{gpt2,fewshot} showed that both zero-shot and few-shot prompting can achieve performance comparable to fine-tuning without further training. Building on these studies~\cite{gpt2}, VideoA11y employs zero-shot prompt engineering to generate descriptions that adhere to AD guidelines and exceed the quality of human-generated descriptions.

\subsection{Video Description Datasets}

Numerous video description datasets have been introduced across various domains, including cooking~\cite{youcook2}, movies~\citep{mad, movienet, cmd}, social media~\cite{tvc}, and human activities~\citep{activity, msvd, vatex, charades, charadesego, videoxum}. Other datasets cover a broader range of video categories~\citep{chen2023valor, msr-vtt, webvid, internvid, panda70m, youkumplug, vitt}. These datasets often include annotations from novice human describers recruited through online platforms, such as MTurk~\citep{tvc, activity, msvd, vatex, msr-vtt}. These human annotations can be brief, incomplete, and prone to spelling and grammar errors, especially when provided by inexperienced annotators~\cite{klie2023annotation,chen2022msrvideo}. %Common errors include duplicate sentences, spelling mistakes, and syntax errors~\cite{chen2022msrvideo}. 
These issues affect the overall quality and usability of the dataset. VideoA11y addresses these issues by automatically generating accurate, grammatically correct descriptions from scratch while also correcting errors in human annotations found in existing video datasets and eliminating bias introduced by different annotators.

Recent datasets developed for video description have also used GPT as a part of their pipelines to aid in data generation~\cite{videocc,internvid,VideoChatGPT,panda70m}. For example, VIDEOCC3M~\cite{videocc} leverages GPT-2~\cite{gpt2} as a decoder, utilizing features encoded by BERT~\cite{bert} to generate video descriptions. 
InternVid~\cite{internvid} and Video-ChatGPT~\cite{VideoChatGPT} apply BLIP2~\cite{blip2} and Tag2Text~\cite{tag2text} to generate initial captions and synthesize them into video descriptions using Vicuna~\cite{vicuna} or GPT-3.5. Meanwhile, Panda-70M~\cite{panda70m} curates 3.8 million high-resolution videos from HD-VILA-100M~\cite{hdvila} uses cross-modality teacher models for captioning, followed by fine-tuning a retrieval model on a selected subset to choose the best caption per video.
Finally, OSCaR~\cite{oscar} uses GPT-4V to create a dataset and benchmark for object state and state change description. 
While these approaches generate descriptions and provide benchmarks, they do not allow for tailoring the descriptions to the needs of BLV users, which is the focus of our work.

\subsection{Evaluation Metrics for Video Descriptions}
Evaluating video descriptions is essential for ensuring their quality and usability across various applications. Most video description evaluations rely on automated metrics, which are efficient, objective, and reproducible~\cite{aafaq2019video}. 

These metrics fall into two categories: n-gram-based and content-based. N-gram-based metrics like BLEU~\cite{bleu}, METEOR~\cite{meteor}, and CIDEr~\cite{cider} measure n-gram overlap between generated and reference descriptions, focusing on precision and recall. Content-based metrics, such as SPICE~\cite{spice}, use scene graphs to compare objects, attributes, and relationships for semantic comparisons. However, these metrics often cannot fully capture the accessibility needs of BLV users.

Research involving BLV users frequently employs subjective evaluation methods. Yet, to our knowledge, no standard user-based metric exists for evaluating video descriptions. %Existing subjective evaluation approaches vary widely in scope-some are overly general, while others are highly detailed. 
Existing work often asks participants to give an overall rating~\cite{shortscribe} for video descriptions or rate custom statements (e.g., ``It provides me with useful additional information about the video'')~\cite{spica}. Recently, Natalie et al.\ created a qualitative codebook about different aspects of video description to guide novice describers in evaluating descriptions~\cite{metrics}. 
%For example, ShortScribe~\cite{shortscribe} asked evaluators to assign an overall score to each video description, whereas the SPICA~\cite{spica} had evaluators rate descriptions based on several specific questions. To improve the efficiency and standardization of subjective evaluations, 
We build on this codebook by proposing four custom metrics tailored to BLV users' needs: descriptive, objective, accurate, and clear. These metrics provide a structured framework to guide human evaluators and ensure consistency in assessing video descriptions.

\section{Overview of Our Process and Evaluation Metrics}

\noindent \textbf{Our Process.} We developed and evaluated VideoA11y (method) and VideoA11y-40K (dataset) in four steps:

\begin{enumerate}
\item\textbf{Developing the method: VideoA11y (Section~\ref{sec:method}):} We compiled 42 AD guidelines from online sources for professional describers and designed a compliant prompt based on these guidelines. To select an MLLM for VideoA11y, we applied our prompts to SOTA open-source model Video-LLaVA~\cite{videollava} and proprietary model GPT-4V~\cite{openai2024gpt4} to create descriptions for 150 videos sampled from 15 different categories. Subsequently, we conducted a study with 150 sighted users on MTurk to evaluate the quality of the generated descriptions (Study 1).  
\item\textbf{Creating the dataset: VideoA11y-40K (Section~\ref{sec:datasets}):} Based on Step 1, we selected GPT-4V as the MLLM for VideoA11y to generate descriptions for 40,000 videos, resulting in the VideoA11y-40K dataset.
\item\textbf{Evaluating VideoA11y and VideoA11y-40k with sighted novices, professional describers, and BLV users (Section~\ref{sec:evaluation}):} We conducted four user studies to evaluate the method and dataset. The first two studies involved 197 sighted users: Study 2 compared the quality of video descriptions generated by our method with existing human annotations in video datasets, while Studies 3 and 4 asked sighted novices and professional describers to compare VideoA11y descriptions with high-quality annotations created by trained humans. Study 5 involved 40 BLV users who assessed the descriptions generated by VideoA11y compared to those produced by novice human describers.

\item\textbf{Technical experiments to provide a benchmark for video accessibility (Section~\ref{sec:experiments}):} We fine-tuned two open-source MLLMs, Video-LLaVA~\cite{videollava} and LLaVA-Next-Video~\cite{llavanext}, on VideoA11y-40K. We then evaluated the fine-tuned models, along with baselines and other MLLMs, using a range of standard and custom metrics. This evaluation provides a benchmark for future video description models tailored to the needs of BLV users.

\end{enumerate}

\noindent \textbf{Metrics for Human and Technical Evaluations and Benchmarking.}
\label{sec:metrics}
We employed a combination of custom and standard metrics to evaluate VideoA11y and VideoA11y-40K. For the custom metrics, we identified four specific metrics from the accessibility literature~\cite{metrics} and provided their definitions to participants in our studies. The four custom metrics are \textit{descriptive}, \textit{objective}, \textit{accurate}, and \textit{clear}. The \textit{descriptive} metric evaluates whether the description provides detailed yet concise information about objects, people, and settings. The \textit{objective} metric assesses whether only visible elements are reported without incorporating personal opinions or assumptions. The \textit{accurate} metric focuses on the precision and correctness of details such as colors and spatial arrangements. The \textit{clear} metric examines whether the information is presented in a way that is easy to follow and understand, avoiding confusion. Complete definitions of these metrics are provided in Appendix~\ref{app:metrics}. In addition, we used six standard metrics from the NLP domain: Bleu\_1~\cite{bleu}, Bleu\_4~\cite{bleu}, METEOR~\cite{meteor}, ROUGE\_L~\cite{rouge}, CIDEr~\cite{cider}, SPICE~\cite{spice}. These metrics assess various aspects of quality, including n-gram precision and recall (Bleu, METEOR, ROUGE) and semantic relevance and alignment with human judgment (CIDEr, SPICE).

\section{Developing the Method: VideoA11y}
\label{sec:method}
VideoA11y employs MLLMs and video accessibility guidelines to produce precise and clear descriptions for BLV individuals. This process involves analyzing video frames and, when available, incorporating existing human annotations. We compiled AD guidelines from accessibility resources (Section~\ref{sec:ad}), which were then integrated into a carefully crafted prompt. This compliant prompt, along with the keyframes (Section~\ref{sec:keyframes}), is passed to an MLLM to generate or revise video descriptions (Section~\ref{sec:prompt}).

\begin{figure*}
    \centering
    \includegraphics[width=0.80\linewidth]{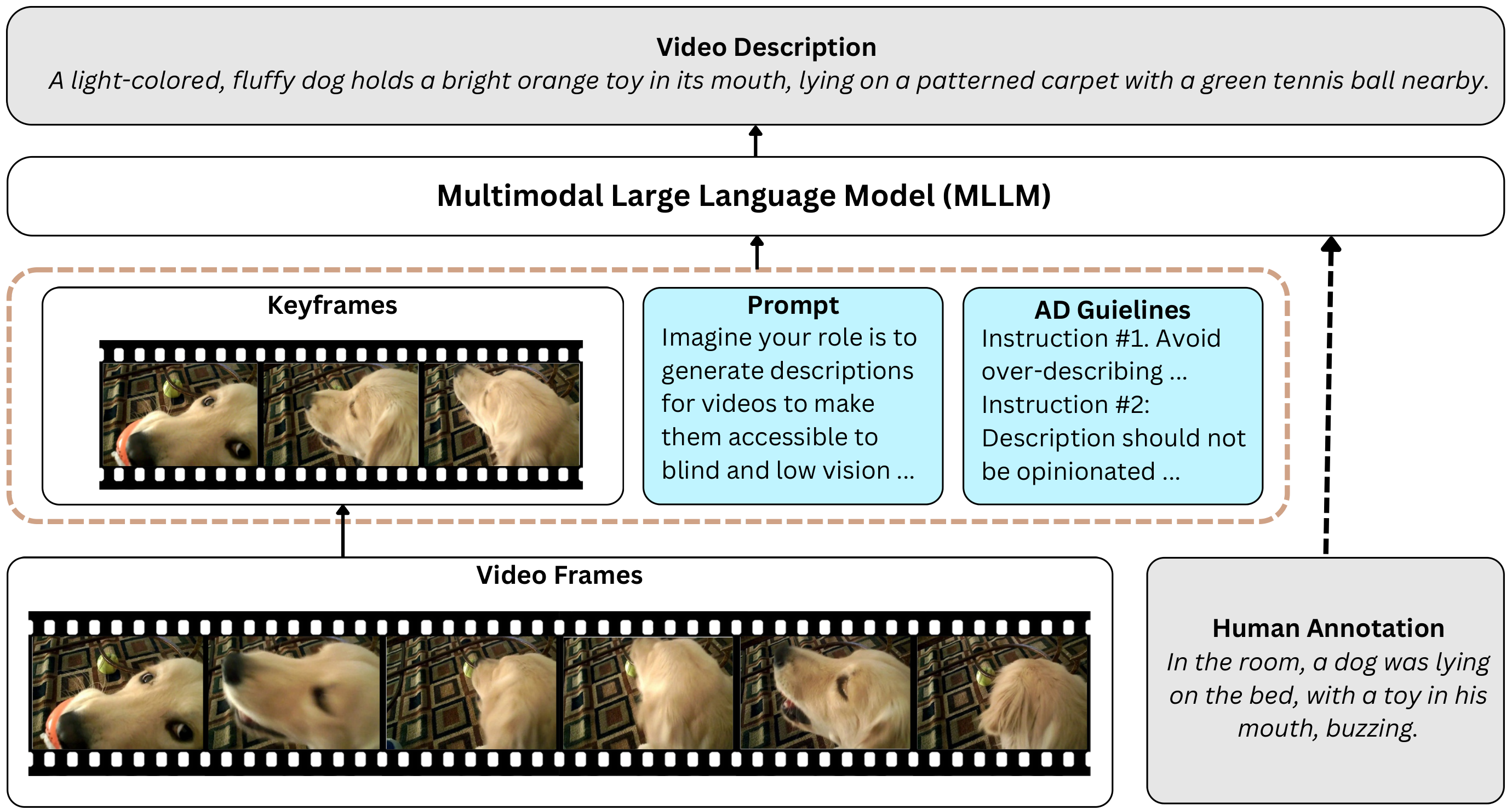}
    \caption{Overview of the VideoA11y pipeline. First, keyframes are extracted from the input video. Then, the keyframes, the prompt, AD guidelines, and optional human annotations are provided to MLLM, which generates accessible video descriptions.}
    \label{method}
    \Description{The figure provides an overview of the VideoA11y pipeline, illustrating how video descriptions are generated. The process starts with extracting keyframes from input video frames. These keyframes, along with a descriptive prompt, AD (Audio Description) guidelines, and optional human annotations, are provided to the MLLM. The MLLM then generates a detailed and accessible video description, incorporating elements from the keyframes, prompt, and guidelines.}
\end{figure*}

\subsection{Curating Audio Description Guidelines}
\label{sec:ad}
We initially collected a total of 154 AD guidelines from four different online sources: Netflix Accessibility Guidelines~\cite{netflix}, Ofcom Guidelines~\cite{ofcom2021}, Media Access Canada Guidelines~\cite{canada}, and the Described and Captioned Media Program (DCMP)~\cite{dcmp}. 
These guidelines are curated for professional audio describers but have also been used to train novice describers to create descriptions for videos. They cover general guidelines for creating audio descriptions~\cite{ofcom2021}, as well as guidelines specific to educational~\cite{dcmp} and entertainment content~\citep{netflix,canada}. While this list of 154 guidelines may not be exhaustive, they capture the majority of instructions agreed upon by professional describers. This was also evident from the overlap of several guidelines across these four resources. The overlapping and repeated guidelines were removed, and then the remainder were categorized based on whether an MLLM can be prompted to generate a description adhering to the guideline or not. This process removed guidelines focused on context, such as \textit{``Description should include known relationships when they have been revealed.''}~\cite{netflix}, as well as those focused on voicing and audio of the video (e.g., \textit{``Describe the source of sounds that may not be immediately recognizable within the video but are pertinent to understanding and appreciation of the content.''}). Next, we shortened some of the guidelines for prompting. For example, \textit{``Avoid over-describing — do not include visual images that are not vital to the understanding or enjoyment of the scene.''}, became \textit{``Avoid over-describing — Do not include non-essential visual details.''}  The result of this process was 42 AD guidelines optimized for prompting MLLMs (Appendix~\ref{app:ad}). 

\subsection{Keyframe Extraction}
\label{sec:keyframes}
Keyframes in a video capture significant changes or transitions within a scene, often representing shifts in content or visual focus. To extract keyframes from an input video, we implemented the local maximum algorithm ~\citep{keyframe, keyframe2} which converts the frames from RGB to LUV color space to focus on luminance, then calculates the absolute difference between successive frames to measure the extent of change between frames. To reduce noise in the frame difference calculations, we applied a smoothing technique, which averages the values over a sliding window of 15 frames. This helps to smooth out minor fluctuations and highlight meaningful changes. Then, we identify peaks or local maxima, i.e., frames where the value is higher than the frames immediately before and after. These peaks represent significant changes in the video, indicating keyframes that likely correspond to scene transitions or important actions.

\enlargethispage{1pt}
\subsection{Prompt Design and Video Description Generation}
\label{sec:prompt}

We created a prompt using the AD guidelines:

\setlength{\fboxsep}{8pt} 
\setlength{\fboxrule}{0.5mm}
\vspace{0.2cm}
\noindent
\fcolorbox{black}{lightgray}{ 
    \begin{minipage}{0.90\linewidth}
    \raggedright
    Imagine your role is to generate descriptions for videos to make them accessible to blind and low vision individuals. You will watch a sequence of keyframes from a video and read the current description of this video. Your task is to revise the current description. Output your result in a dictionary format: \{``Video\_Category'': A string representing the category of video you believe it to be, ``Revised\_Desc'': A string of revised description.\}
    
    ~\\
    Current Description: \{desc\_current\}

    ~\\
    Instructions:
    
    Instruction \#1: Avoid over-describing — Do not include non-essential visual details.
    
    Instruction \#2: Description should not be opinionated unless content demands it.
    
    % Instruction \#3: Choose level of detail based on plot relevance when describing scenes.
    
    % Instruction \#4: Description should be informative and conversational, in present tense and third-person omniscient.

    Instruction \#3: ...
    \end{minipage}
}

\begin{figure*}[t!]
    \centering
    \includegraphics[width=0.8\linewidth]{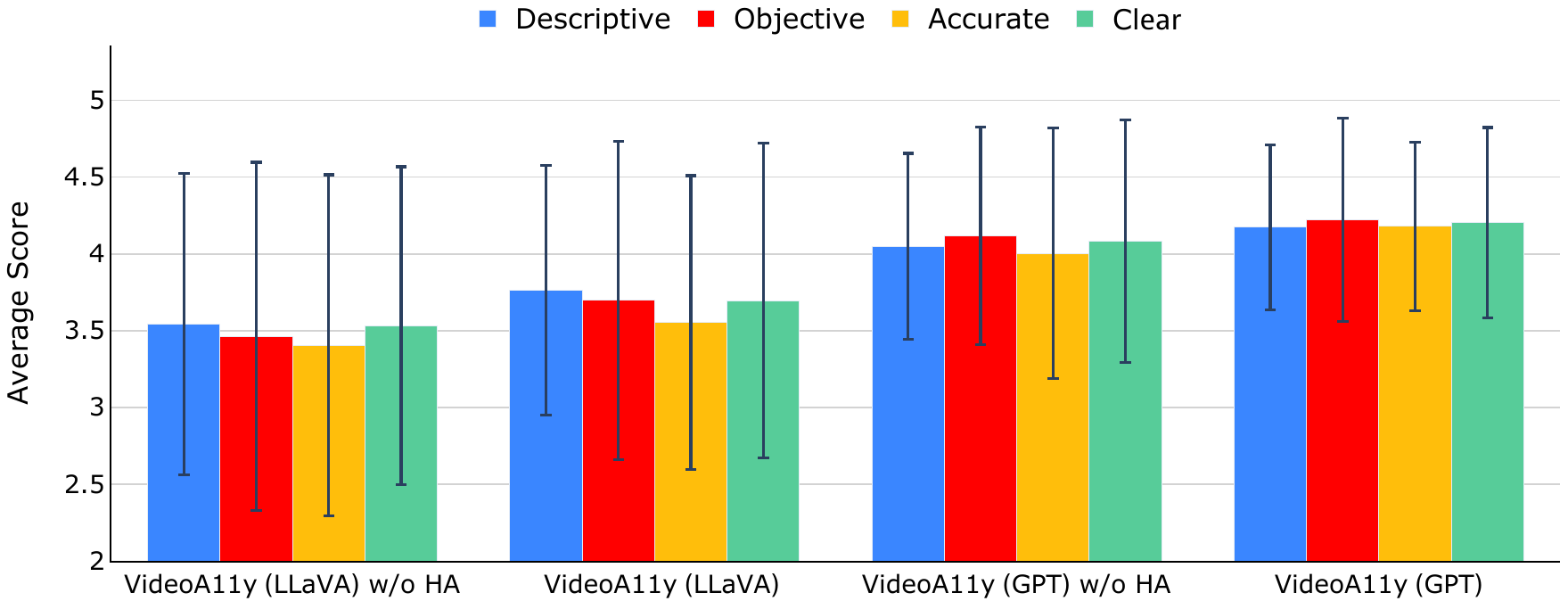}
    \caption{Results of Study 1 with 150 sighted MTurk users. VideoA11y (GPT) outperforms other methods on all metrics ($p<0.05$), followed by VideoA11y (GPT) w/o HA. HA: Human Annotation.}
    \label{Overall1}
    \Description{The bar graph displays average scores for four different video description methods—VideoA11y (LLaVA) without Human Annotation (HA), VideoA11y (LLaVA) with HA, VideoA11y (GPT) without HA, and VideoA11y (GPT)—across four evaluation metrics: Descriptive, Objective, Accurate, and Clear. The scores range from 2 to 5 on the Y axis. VideoA11y (GPT) achieved the highest scores in each metric. Specifically, the scores for VideoA11y (GPT) are approximately 4.2 for Descriptive, 4.2 for Objective, 4.1 for Accurate, and 4.2 for Clear.}
\end{figure*}

\vspace{0.2cm}
We input the extracted keyframes and our compliant prompt, which includes the AD guidelines and optional human annotations, into an MLLM to generate or revise descriptions.

\subsection{Study 1: Evaluating VideoA11y on Open-Source and Proprietary MLLMs}
\label{sec:study1}
We ran an MTurk experiment with sighted users to evaluate the difference in the quality of video descriptions generated by VideoA11y when using open-source vs.\ proprietary MLLMs.
 Specifically, we collected 150 videos sourced from three existing datasets: VALOR32K~\cite{chen2023valor}, VATEX~\cite{vatex}, and YouCook2~\cite{youcook2} to run this evaluation.
We used Video-LLaVA~\cite{videollava} as an open-source MLLM and GPT-4V~\cite{openai2024gpt4} as a proprietary MLLM to generate descriptions, resulting in four conditions:

\begin{enumerate}
    \item \textbf{VideoA11y (LLaVA) w/o HA} uses the compliant prompt and Video-LLaVA to generate descriptions.

    \item \textbf{VideoA11y (LLaVA)} uses the compliant prompt with human annotations and Video-LLaVA to generate descriptions.
    
    \item \textbf{VideoA11y (GPT) w/o HA} uses the compliant prompt and GPT-4V to generate descriptions.

    \item \textbf{VideoA11y (GPT)} uses the compliant prompt with human annotations and GPT-4V to generate descriptions.

\end{enumerate}

We recruited 150 MTurk participants for the study. The participants (74 males, 75 females, 1 preferred not to say) were between 23 and 60 years old and were primarily located in the United States. Each participant watched two videos and read four descriptions for each video. They then rated each description on a 5-point scale of extremely bad, somewhat bad, neither good nor bad, somewhat good, and extremely good for each of the four metrics. The user interface used in Study 1 is shown in Appendix~\ref{app:interface}, and the full list of prompts for the four conditions is shown in Appendix~\ref{app:prompt}.

\subsubsection{Study 1 Results}
Figure~\ref{Overall1} shows the average ratings for the four methods. We used the Friedman Test to analyze our data since the dependent variables (i.e., user ratings) are ordinal. The test reveals a significant effect of the description method. Pairwise comparisons (Appendix~\ref{app:stat_study1}) indicate that VideoA11y (GPT) w/o HA and VideoA11y (GPT) significantly outperform both VideoA11y (LLaVA) w/o HA and VideoA11y (LLaVA) in all four metrics ($p<0.05$). The results also suggest that using human annotations as references can enhance the quality of descriptions, although not significantly ($p>0.05$). Based on these results, we selected GPT-4V as the MLLM and incorporated the existing human annotations in creating the VideoA11y-40K dataset.

\section{Creating the Dataset: VideoA11y-40K}
\label{sec:datasets}
We employed VideoA11y (GPT) to generate high-quality video descriptions for three popular datasets in the computer vision community: VALOR32K~\cite{chen2023valor} (29,635 videos), VATEX~\cite{vatex} (8,765 videos), and YouCook2~\cite{youcook2} (1,600 videos), in accordance with their MIT license permissions. This process resulted in the creation of the VideoA11y-40K dataset, which includes descriptions for 40,000 videos (32,000 training, 4,000 validation, and 4,000 test sets) across 15 categories tailored to BLV users. 

\begin{figure*}[t!]
   \centering
   \begin{subfigure}{0.55\textwidth}
     \centering
     \includegraphics[width=1\linewidth]{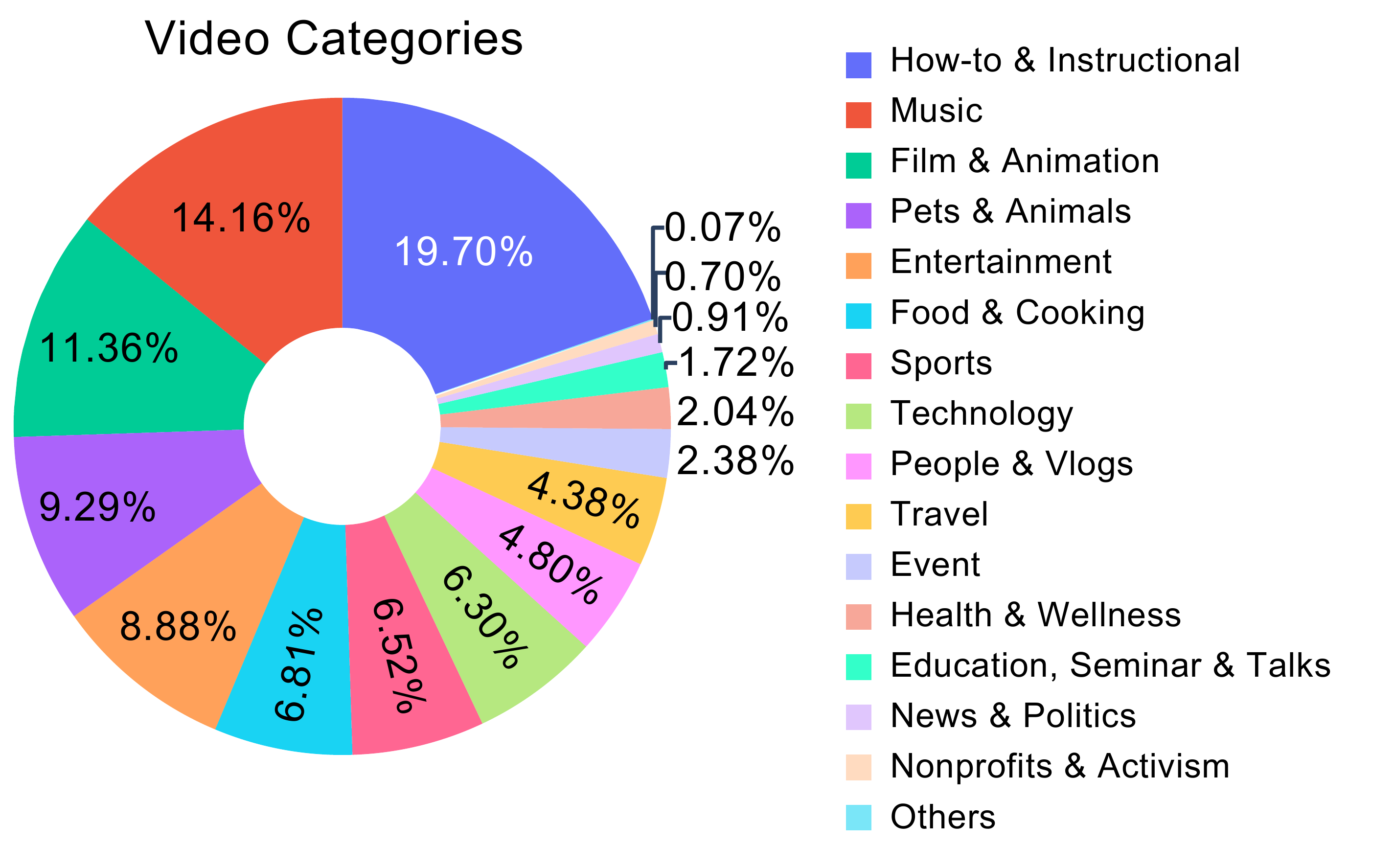} 
     \caption{Distribution of video categories in the VideoA11y-40K dataset.}
     \label{sub:categories}
   \end{subfigure}\hfill
   \begin{subfigure}{0.43\textwidth}
     \centering
     \raisebox{4.5mm}
     {\includegraphics[width=1\linewidth]{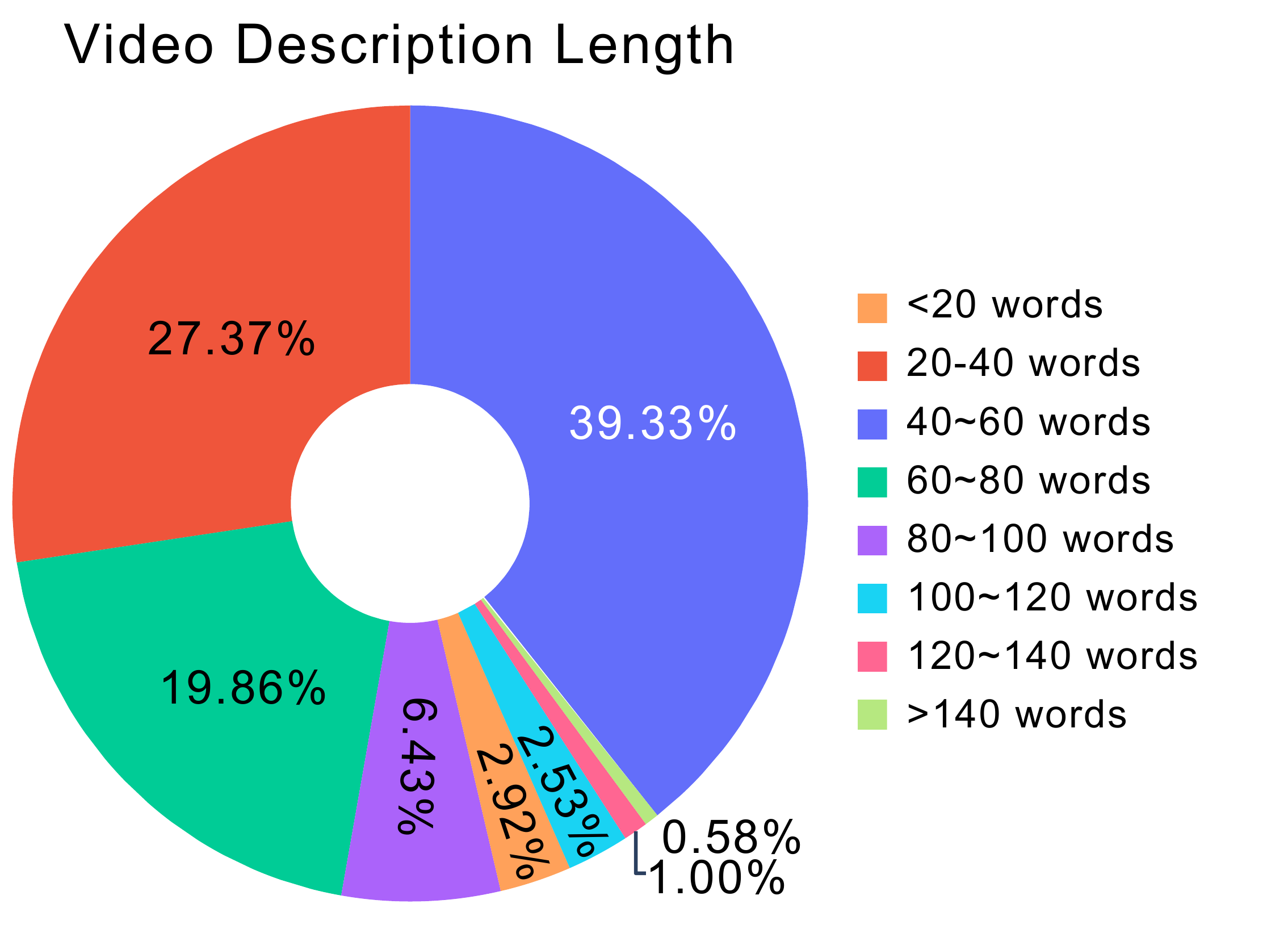}} 
     \caption{Proportions of video descriptions by word count.}
     \label{sub:length}
   \end{subfigure}\hfill
   \caption{Overview of video categories and description lengths in VideoA11y-40K.}
   \Description{The figure displays two pie charts illustrating the distribution of video categories and video description lengths within the dataset. The first chart highlights the variety of video categories, with How-to and Instructional videos making up 19.70\% of the content, followed by Music at 14.16\%, and Film and Animation at 11.36\%. Other notable categories include Pets and Animals (9.29\%), Entertainment (8.88\%), Food and Cooking (6.81\%), Sports (6.52\%), and Technology (6.30\%), among others, with smaller percentages allocated to categories such as People and Vlogs (4.80\%), Travel (4.38\%), Event (2.38\%), Health and Wellness (2.04\%), Education, Seminar and Talks (1.72\%), News and Politics (0.91\%), and Nonprofits and Activism (0.70\%). The second chart focuses on video description lengths, revealing that 39.33\% of descriptions fall within the 40-60 words range, 27.37\% are 20-40 words long, 19.86\% contain 60-80 words, and 6.43\% contain 80-100 words. Shorter and longer descriptions are less common, with descriptions under 20 words and over 100 words making up 2.92\% and 5.11\% of the dataset, respectively.}
   \label{fig:video_stats}
\end{figure*}

\subsection{Video Categorization}
\label{sec:dataset_construction}

We derived 15 video categories by adapting and merging existing categories from YouTube to ensure comprehensive coverage. The categories are:
(1) Film and Animation; (2) Music; (3) Sports; (4) Entertainment; (5) News and Politics; (6) Pets and Animals; (7) How-to and Instructional; (8) Event; (9) Travel, (10) People and Vlogs; (11) Food and Cooking; (12) Health and Wellness; (13) Auto and Technology, (14) Nonprofits and Activism; and (15) Education, Seminar and Talks. Each video in VideoA11y-40K was assigned to one of these 15 categories by using GPT-4 to analyze the video descriptions generated by VideoA11y. To verify categorization accuracy, we randomly sampled 5 videos from each category (75 videos in total) and recruited 225 MTurk participants (152 males, 71 females, aged 21--68 years) to rate the correctness of the assigned categories. A video was deemed misclassified if at least two out of three votes indicated an incorrect category. Our results confirmed that 96\% (72 out of 75) of the videos were accurately categorized.

\subsection{Dataset Statistics}
\label{sec:dataset_statistics}
The average description length in the VideoA11y-40K dataset is $52.30$ words, which is considerably longer than $20.30$ words in the original datasets. Figure~\ref{sub:length} presented the description length distribution in VideoA11y-40K, with the majority of captions ranging between 40-60 words (39.33\%) and 20-40 words (27.37\%). Descriptions shorter than 20 words or longer than 140 words comprise only 2.92\% and 0.58\%, respectively. Figure~\ref{sub:categories} showed video distribution across 15 categories; How-to and Instructional (19.7\%), Music (14.16\%), and Film and Animation (11.36\%) dominate, while News and Politics and Nonprofits and Activism each make up less than 1\% of the videos.

\section{Evaluating VideoA11y Method and VideoA11y-40K Dataset with Sighted and BLV Users}
\label{sec:evaluation}
We evaluated VideoA11y and VideoA11y-40K through human subject studies with sighted and BLV individuals. Specifically, we sampled 150 videos from VideoA11y-40K, evenly distributed across the 15 categories introduced in Section~\ref{sec:dataset_construction} (10 videos per category). 
We conducted two studies with sighted users on MTurk (Sections~\ref{sec:study2}, Section~\ref{sec:study3}), one online study with professional describers (Section~\ref{sec:studypd}), and an online study with BLV participants (Section~\ref{sec:studyblv}) to evaluate the quality of the video descriptions in the dataset. We obtained IRB approval for all studies. Participants viewed the informed consent sheet on the first page of the online surveys and provided consent by proceeding. For these studies, we set the statistical significance level at $p=0.05$. The full statistical results are presented in Appendix~\ref{app:stat}.

\subsection{Study 2: Comparison with Descriptions Produced by Novice Humans}

\label{sec:study2}
We ran an MTurk experiment with sighted users to assess the quality of the descriptions in VideoA11y-40K compared to descriptions created by novice annotators in the original datasets. We also included GPT-4V-generated descriptions that did not use the AD guidelines in the prompt (i.e., non-compliant prompt) to assess how they compare to novice human annotations. We recruited 150 new MTurk participants for the study. The participants (98 males and 52 females) were between 22 and 60 years old. Of these, 149 were from the United States, and one was from Italy.  Each participant watched two videos and rated the following five descriptions for each video:

\begin{enumerate}
    \item \textbf{Human Annotation} uses novice human annotations from the original datasets.
    \item \textbf{GPT-4V} uses the non-compliant prompt to generate descriptions. 
    \item \textbf{GPT-4V w/ HA} uses the non-compliant prompt with human annotations to generate descriptions. 
    \item \textbf{VideoA11y w/o HA} uses the compliant prompt to generate descriptions.

    \item \textbf{VideoA11y} uses the compliant prompt with human annotations to generate descriptions.
   
\end{enumerate}

The user interface used in Study 2 and the list of prompts are shown in Appendices~\ref{app:interface} and~\ref{app:prompt}, respectively.

\begin{figure*}
    \centering
    \includegraphics[width=0.8\linewidth]{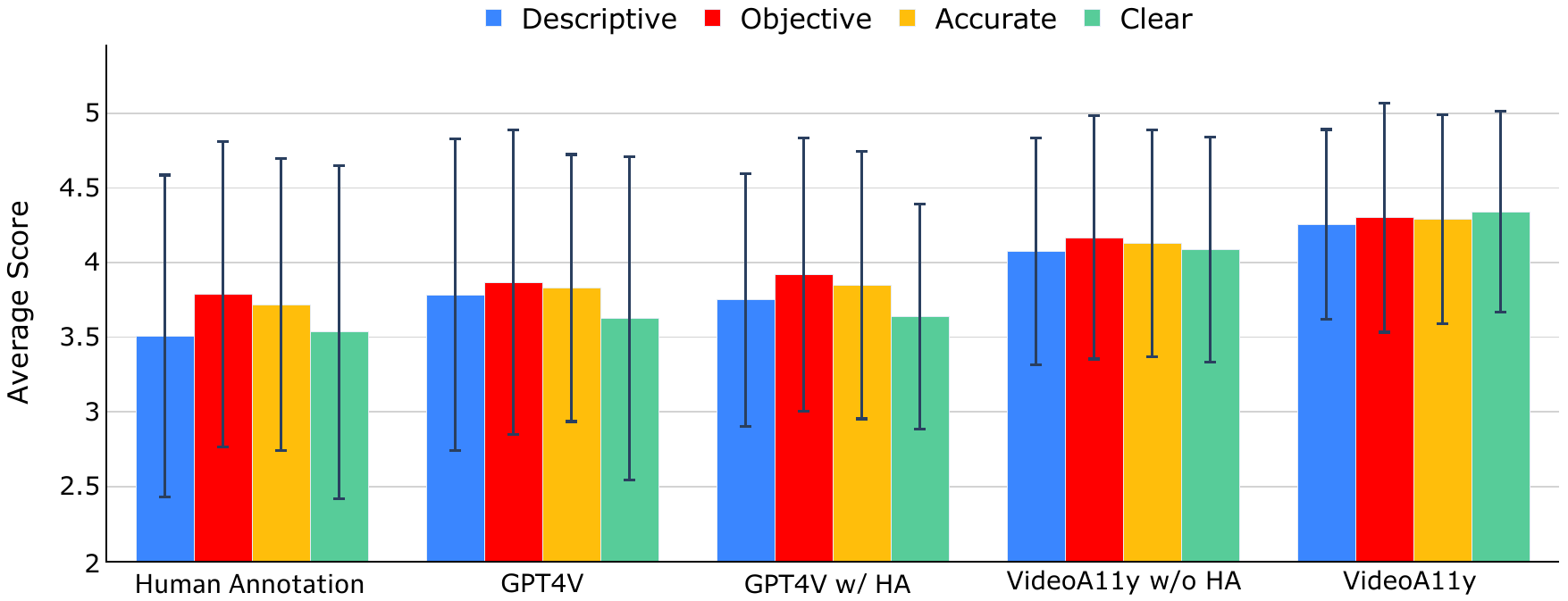}
    \caption{Results of Study 2 with 150 sighted MTurk users. VideoA11y outperforms other methods in all metrics ($p<0.001$), followed by VideoA11y w/o HA. HA: Human Annotation.}
    \label{Overall2}
    \Description{The bar graph displays average scores for five different video description methods—Human Annotation, GPT-4V, GPT-4V with Human Annotation (HA), VideoA11y without HA, and VideoA11y—across four evaluation metrics: Descriptive, Objective, Accurate, and Clear. The scores range from 2 to 5 on the Y axis. VideoA11y achieved the highest scores in each metric. Specifically, the scores for VideoA11y are approximately 4.2 for Descriptive, Objective, and Accurate, and 4.4 for Clear. Human Annotation, on the other hand, scored lower in all metrics, with scores around 3.5 for Descriptive, 3.7 for Objective, 3.7 for Accurate, and 3.6 for Clear.}
\end{figure*}

\subsubsection{Study 2 Results}
Figure \ref{Overall2} shows the average ratings for the five methods. As in Study 1, we used the Friedman Test to analyze the ratings. The pairwise comparisons, with significance values for multiple comparisons (Appendix~\ref{app:stat_study2}), show that VideoA11y offers significant enhancements in all four metrics compared to Human Annotation, GPT-4V, and GPT-4V w/ HA, with all comparisons resulting in $p$-values under $0.001$, suggesting the effectiveness of our approach in enhancing video description quality. In addition, VideoA11y w/o HA demonstrates statistically significant improvements over Human Annotation, GPT-4V, and GPT-4V w/ HA in all four metrics, with all $p$-values below $0.02$. Finally, the overall performance of baseline GPT-4V and GPT-4V w/ HA is comparable to novice human annotations ($p>0.05$) across all metrics, except for the `descriptive' metric where baseline GPT-4V shows a significantly better performance with $p<0.05$. These results indicate the effectiveness of AD guidelines in improving description quality beyond novice human annotations. While these results highlight the strengths of VideoA11y, minor inaccuracies can still occur in certain cases, as illustrated with examples in Appendix~\ref{app:qualitative}.

\subsection{Study 3: Comparison with Descriptions Produced by Trained Humans}
\label{sec:study3}

\begin{figure*}[t!]
   \raggedright
   \begin{subfigure}{0.45\textwidth}
     \centering
     \includegraphics[width=1\linewidth]{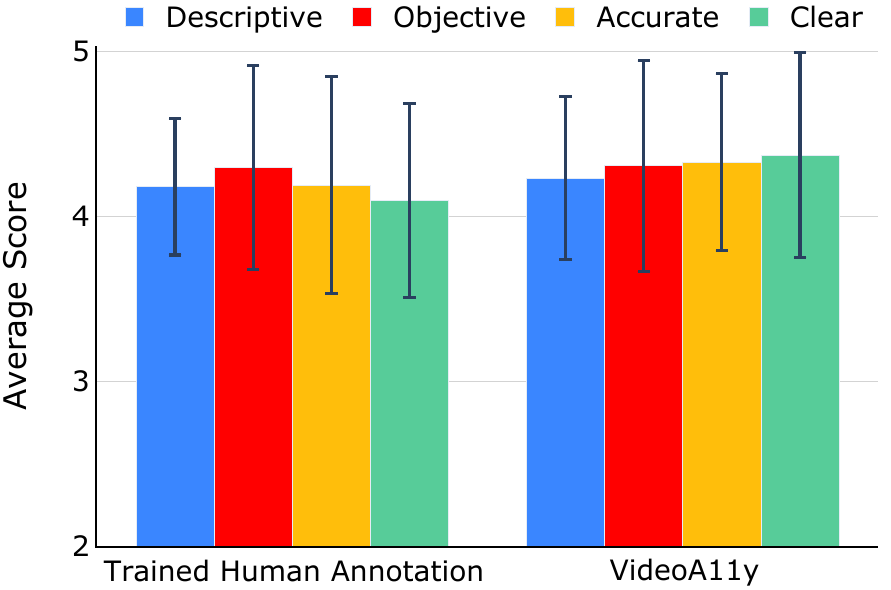} 
     \caption{Results of Study 3 with 47 sighted MTurk users. VideoA11y received higher average ratings than trained humans on all metrics, with a statistically significant difference on the clear metric ($p = 0.004$).}
     \label{sub:study3}
   \end{subfigure}
   \hspace{0.05\textwidth}
   \begin{subfigure}{0.45\textwidth}
     \centering
     {\includegraphics[width=1\linewidth]{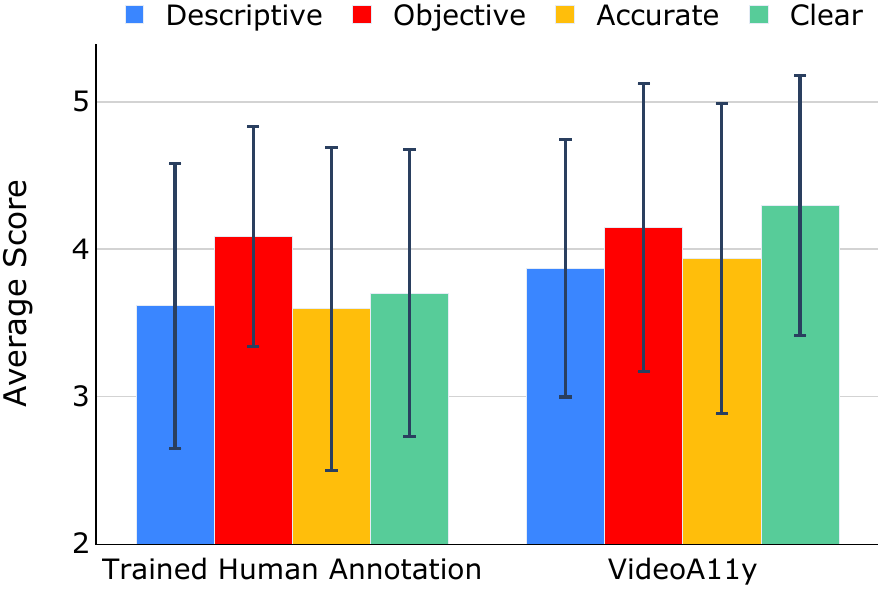}} 
     \caption{Results of Study 4 with seven professional describers. VideoA11y received higher average ratings than trained humans on all metrics, although the differences were not statistically significant ($p > 0.05$).}
     \label{sub:study4}
   \end{subfigure}
   \caption{Comparison of VideoA11y's descriptions and trained human annotations on the 47 videos, evaluated by sighted MTurk users and professional describers.}
   \label{fig:high_ablation}
   \Description{The figure presents two subfigures comparing the performance of VideoA11y and trained human annotations across four evaluation metrics: Descriptive, Objective, Accurate, and Clear. Subfigure (a) shows the results from Study 3, where 47 sighted MTurk participants evaluated descriptions generated by VideoA11y and trained humans. The vertical bar chart demonstrates that VideoA11y outperformed trained humans in all four metrics, with approximate scores of 4.2 for Descriptive, 4.3 for Objective, 4.3 for Accurate, and 4.4 for Clear, compared to Human Annotation's scores of around 4.2, 4.3, 4.2, and 4.1, respectively. Subfigure (b) displays the results from Study 4, where seven professional audio describers assessed the same descriptions. The vertical bar chart indicates that VideoA11y outperformed trained humans in all four metrics, with approximate scores of 3.9 for Descriptive, 4.1 for Objective, 4.0 for Accurate, and 4.3 for Clear, compared to Human Annotation's scores of around 3.6, 4.1, 3.6, and 3.7, respectively.}
\end{figure*}

In this study, we evaluated whether the descriptions generated by VideoA11y can meet the same standards as those produced by human describers who carefully follow AD guidelines. 
We selected 47 videos (ranging from 4 to 7 minutes long, with an average length of 4.92 minutes) from YouTube across various categories, and our team of four accessibility researchers created descriptions for these videos in accordance with the 42 AD guidelines we curated. We aimed to ensure that the description quality adheres to the standards and guidelines set by professional audio describers. We then recruited 47 sighted participants via MTurk to evaluate the descriptions generated from VideoA11y and those created by trained humans. In this study, we used GPT-4V as the MLLM for VideoA11y. On average, VideoA11y’s descriptions (130.51 words) are comparable in length to human descriptions (140.17 words), ensuring a fair comparison.
%Table~\ref{tab:high_quality} 
Figure~\ref{sub:study3} shows that the average ratings for VideoA11y descriptions are higher than those for high-quality human annotations in all four metrics (descriptiveness, objectivity, accuracy, and clarity). Furthermore, the Wilcoxon Signed-Rank test (Appendix~\ref{app:stat_study3}) demonstrates that the descriptions generated by VideoA11y show a statistically significant improvement ($p<0.05$) in the ‘clear’ metric compared to high-quality human annotations. Thus, we used VideoA11y descriptions as the ground truth to conduct the additional technical experiments reported below.

% \begin{figure}
%     \centering
%     \includegraphics[width=0.40\linewidth]{assets/high_survey_short.pdf}
%     \caption{Results of Study 3 with 47 sighted MTurk users. VideoA11y outperformed trained humans on the `clear' metric ($p=0.004$) and performed similarly to trained humans on other metrics ($p>0.05$).} %Ablation study results on 47 videos with high-quality human annotations.}
%     \label{fig:high_ablation}
%     \Description{The bar graph displays average scores for two video description methods—Trained Human Annotation and VideoA11y—across four evaluation metrics: Descriptive, Objective, Accurate, and Clear. The scores range from 2 to 5 on the Y axis. VideoA11y outperformed Trained Human Annotation on the 'Clear' metric with a score of approximately 4.5, while Trained Human Annotation scored around 4.2. For the remaining metrics (Descriptive, Objective, and Accurate), both VideoA11y and Trained Human Annotation performed similarly, with scores around 4.1 to 4.3.}
% \end{figure}

\subsection{Study 4: Evaluation with Professional Describers}
\label{sec:studypd}

We conducted a study with seven professional audio describers, each with over three years of paid experience (Appendix~\ref{app:demographic_expert}), to evaluate the quality of VideoA11y's descriptions compared to those created by trained humans.  Finding professional describers is challenging, and prior work on video accessibility included one to three professional describers~\citep{kobayashi2009, omniscribe}. In our study, the seven experts viewed the 47 videos from Section~\ref{sec:study3}, rated both sets of descriptions on four metrics, and selected the better description for each video with reasoning. Additionally, they participated in a Zoom interview to provide qualitative insights.
Figure~\ref{sub:study4} shows that the average expert ratings for VideoA11y descriptions exceed those for trained human annotations on all four metrics. Also, the differences in ratings between VideoA11y and trained humans are more noticeable when evaluated by expert describers. A Wilcoxon Signed-Rank test (Appendix~\ref{app:stat_study4}) reveals no statistically significant differences ($p>0.05$) on all four metrics between VideoA11y and human descriptions, likely due to the small sample size. The medium to large effect sizes for three metrics ($0.459$--$0.640$) suggest the difference in the ratings is important. Additionally, professional audio describers preferred VideoA11y's descriptions for 33 (out of 47) videos. 

While unaware of the description source, the experts liked the choice of words, sentence structure, and visual details in VideoA11y’s descriptions, noting these as important factors in conveying the visual feel and narrative intent of the videos ($n=4$). They found VideoA11y's descriptions more accurate in describing the events ($n=4$) and actions and described them as ``engaging (P1)’’, ``flavorful (P3)’’, and with good flow between sentences. 
Regarding visual detail, the experts noted that, overall, with a comparable length to human annotations, VideoA11y's descriptions included more information on character appearance, reactions, environment, and on-screen text compared to human descriptions. They suggested to include even more details on \textit{``gender identity, race, approximate age, range, and body composition''} (P3). 

In contrast, they stated the human annotations were objective, but in some cases ``too literal (P5)’’ and ``bland'' (n=3): \textit{``Sometimes it was the sentence structure, it was so boring and not artistic. I would choose the one that varied in pacing. (P1)''} 
Among the 14 videos with preferred human annotation, seven videos had notably longer descriptions than VideoA11y’s. These descriptions included details on the physical appearance of people, scenes, and object features (e.g., a watch, types of rice). Also, humans better narrated the purpose of actions in two videos. % (x2), was mentioned as slighty more descriptive (x2) and with better transition between sentences (x1). 
These comments suggest that, in some cases, trained humans could surpass AI quality in providing details.

When asked if they could identify VideoA11y's descriptions from the human descriptions, participants could not distinguish between them ($n=3$) or wrongly identified the human descriptions as AI-generated ($n=4$). The experts thought the more objective and literal descriptions were AI-generated (which was not the case). P5 justified: \textit{``AI is really good at collecting information, but AI isn’t great at integrating or resynthesizing it into a more human voice... Description 1 [VideoA11y's] tended to be more like how I would write them.''} P4 and P7 made similar comments. This feedback highlights VideoA11y's ability to align more closely with established professional standards for description writing.

\subsection{Study 5: Evaluation with Blind and Low Vision Individuals}
\label{sec:studyblv}
We recruited 40 BLV participants to evaluate the effectiveness of VideoA11y and VideoA11y-40K. The participants' demographic information is shown in Appendix~\ref{app:demographic_blv}.
Six participants were completely blind, and 34 were legally blind with a visual acuity ranging from 20/200 to 20/1000~\cite{dandona_2006_lowvision}. We selected 10 videos (2 per category) from (1) Entertainment, (2) How-to and Instructional, (3) Sports, (4) Pets and Animals, and (5) People and Vlogs in VideoA11y-40K dataset. We divided the participants into two groups of 20, with each group evaluating five videos. We inserted human annotations from the original datasets and VideoA11y-40K descriptions as audio descriptions at timestamps preceding the video segments they referred to according to the AD best practices~\cite{dcmp,canada}. After reading the definition of the four evaluation metrics, BLV participants watched each video once with existing human descriptions and again with VideoA11y-40K descriptions with counterbalanced presentation order and rated the descriptions on four metrics of descriptiveness, objectiveness, accuracy, and clarity using a 10-point Likert scale. They also selected which description they preferred for each video and provided reasons for their selection without knowing the description source. We show the user interface in Appendix~\ref{app:interface}.

\subsubsection{Study 5 Results}
\begin{figure*}[t!]
   \centering
   \begin{subfigure}{0.535\textwidth}
     \centering
     \includegraphics[width=1\linewidth]{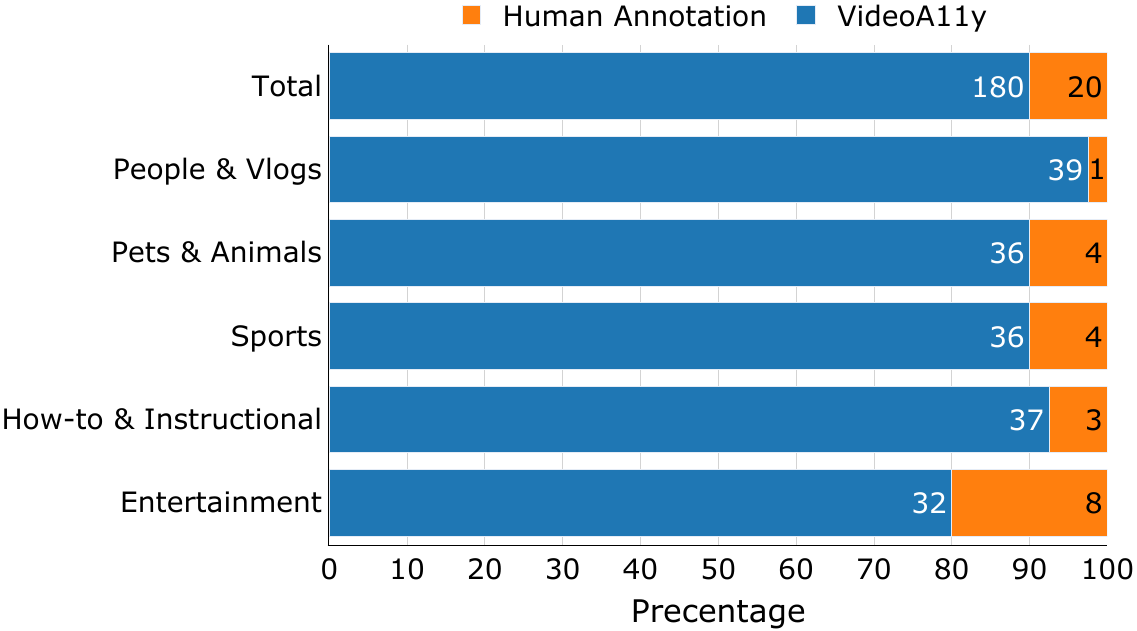}
     \caption{VideoA11y outperforms novice human annotations in all five video categories ($p<0.05$).}
     \label{sub:select}
   \end{subfigure}\hfill
   \begin{subfigure}{0.445\textwidth}
     \centering
     \includegraphics[width=1\linewidth]{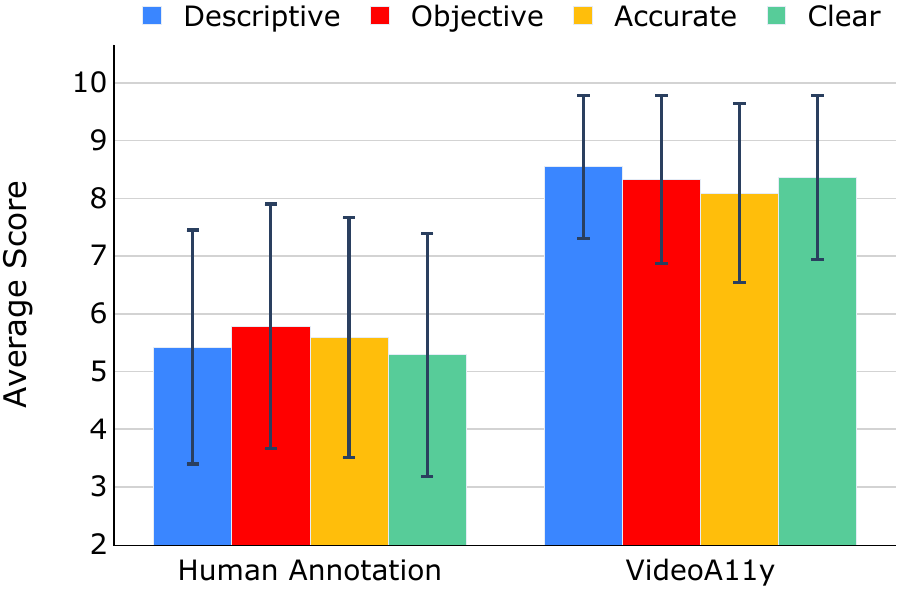} 
     \caption{VideoA11y outperforms novice human annotations in all four metrics ($p<0.001)$.}
     \label{sub:plot}
   \end{subfigure}\hfill
   \caption{Results of Study 5 with 40 blind and low vision users.}
   \label{fig:blind_survey}
   \Description{The figure presents two subfigures comparing the performance of VideoA11y and Human Annotation across various video categories and evaluation metrics. Subfigure (a) shows a horizontal bar chart illustrating the percentage preference for VideoA11y and Human Annotation across five video categories: People and Vlogs, Pets and Animals, Sports, How-to and Instructional, and Entertainment, as well as the total preference. VideoA11y is preferred significantly more in all categories, with total preferences showing 90\% for VideoA11y and 10\% for Human Annotation. Specifically, VideoA11y receives 97.5\% preference in People and Vlogs, 90\% in Pets and Animals, 90\% in Sports, 92.5\% in How-to and Instructional, and 80\% in Entertainment, compared to Human Annotation's 2.5\%, 10\%, 10\%, 7.5\%, and 20\%, respectively. Subfigure (b) presents a vertical bar chart that compares the average scores for four evaluation metrics—Descriptive, Objective, Accurate, and Clear—between VideoA11y and Human Annotation. VideoA11y consistently scores higher in all metrics, with approximate scores of 8.5 for Descriptive, 8.3 for Objective, 8.1 for Accurate, and 8.3 for Clear, compared to Human Annotation's scores of around 5.4, 5.8, 5.6, and 5.3, respectively.}
\end{figure*}

Results from BLV users show that VideoA11y outperforms novice human describers in all five video categories and achieves a selection rate exceeding 80\% in every category (Figure \ref{sub:select}). Videoa11y had a selection rate of 90\% (180 out of 200), demonstrating that our approach significantly enhanced the ability of BLV individuals to understand and enjoy video content. We also compared ratings for the four metrics between human annotations and VideoA11y (Figure \ref{sub:plot}). Based on the Wilcoxon Signed-Rank test results in Table~\ref{tab:blind_pairwise_comparisons}, VideoA11y significantly outperforms human annotations in all metrics, with ratings of $8.54$ vs. $5.43$ (descriptive), $8.33$ vs. $5.79$ (objective), $8.09$ vs. $5.59$ (accurate), and $8.36$ vs. $5.29$ (clear) with all p-values below $0.001$. The results indicate VideoA11y's efficacy in enhancing video understanding for BLV users. 

Comments from BLV users highlighted factors that impacted their description choices and ratings. 28 participants valued the clarity of descriptions generated by VideoA11y, while 25 highlighted the detailed descriptions facilitated their understanding of the videos and the context of the events. 
For example, P16 noted for the Entertainment video:
\textit{``At the end of the video, we had the man crying, the second audio [human annotated] failed to tell us why he was. The first audio [VideoA11y] however told us he did it `playfully'. This is a very vital information as it adds a whole new context to it. The first audio [VideoA11y] also narrated the events accordingly.''}
In addition, 17 participants noted the accuracy and completeness of VideoA11y descriptions in providing details about specific actions and visual elements that were often missing or less comprehensive in the human annotations. P17 highlighted for the Pets and Animals video: 
\textit{``It [VideoA11y] completely describes the actions of the woman from when she walked out with her horse and tied the horse. But the second video [human annotation] didn't talk about what is happening currently and at some points I was lost because I didn't understand what was being said.''}
P36 also mentioned this point for the Entertainment video: \textit{``The second audio description [VideoA11y] was more detailed in term of object, colour, shape,  clothing, the actions in the video, facial expressions of the man while the first [human annotated] seems a bit short of details.''}

\begin{table}[h]
\footnotesize
\centering
\caption{Overall pairwise comparisons from BLV user evaluations between VideoA11y and novice human descriptions in Study 5.}
\label{tab:blind_pairwise_comparisons}
\begin{tabular}{>{\raggedright\arraybackslash}p{0.37\linewidth}
>{\raggedright\arraybackslash}p{0.13\linewidth}
>{\raggedright\arraybackslash}p{0.08\linewidth}
>{\raggedright\arraybackslash}p{0.11\linewidth}
>
{\raggedright\arraybackslash}p{0.10\linewidth}}
\toprule
\textbf{Condition 1 | Condition 2} & \textbf{Metric} & \textbf{Effect Size} & \textbf{Test Statistics} & \textbf{P Value}  \\
\midrule
VideoA11y | Human Annotaion& Descriptive & 0.805 & 11.387  & \textbf{$<$0.001} \\
\midrule
VideoA11y | Human Annotaion& Objective & 0.711 & 10.052 & \textbf{$<$0.001} \\
\midrule
VideoA11y | Human Annotaion& Accurate& 0.708 & 10.014 & \textbf{$<$0.001} \\
\midrule
VideoA11y | Human Annotaion& Clear& 0.772 & 10.920 & \textbf{$<$0.001} \\
\bottomrule
\end{tabular}
\begin{minipage}{0.98\linewidth}
\vspace{0.1cm}
\footnotesize
    Each row tests the null hypothesis that the Condition 1 and Condition 2 distributions are the same. Asymptotic significances (2-sided tests) are displayed. The significance level is 0.05. All effect sizes are large, indicating practical significance.
\end{minipage}
\end{table}

Furthermore, nine BLV participants noted that VideoA11y descriptions closely matched the unfolding events.
For instance, P15 mentioned for the How-to and Instructional video: \textit{``It [VideoA11y] described every action that took place accurately with the audio being in sync with the video.''}
Lastly, eight participants favored descriptions from VideoA11y for providing a more balanced and impartial depiction of events, ensuring that all aspects of the scene were given attention.
 P13 emphasized for the Sports video: 
\textit{``In the first video [VideoA11y], the arm wrestling matches are described in more detail, and both men are assigned a shirt color so that viewers can place each contestant. In the second video [human annotation], the description focuses on the winner, and does not discuss the progression of the matches at all.''}
Overall, these responses underscore that video descriptions generated by VideoA11y enhanced the viewing experience for BLV users by offering clear, detailed,  accurately synchronized, and less biased narratives compared to human annotations.

\section{Technical Experiments to Provide a Benchmark for Video Accessibility} 
\label{sec:experiments}

%We benchmarked the performance of SOTA open-source models as candidates for the MLLM in VideoA11y in generating accessible video descriptions. This involved two complementary evaluations: one benchmarking VideoA11y itself and another comparing models fine-tuned on the VideoA11y-40K dataset. 
We benchmarked the performance of SOTA open-source models as candidates for the MLLM in VideoA11y through two complementary evaluations: (1) benchmarking the VideoA11y method and (2) benchmarking models fine-tuned on the VideoA11y-40K dataset. Both evaluations used standard and custom metrics, with the VideoA11y-40K descriptions serving as the ground truth. This choice is supported by the studies above with sighted and BLV users, which show that the quality of VideoA11y-40K descriptions surpasses even that of trained human annotations. For benchmarking VideoA11y, we assessed its ability to enhance the performance of diverse MLLMs without fine-tuning to highlight its generalizability. For benchmarking VideoA11y-40K, we compared fine-tuned models against SOTA open-source baselines. All evaluations were conducted on the held-out test set of VideoA11y-40K.

%\subsection{Open-Source Models}

\subsection{Benchmarking VideoA11y}
\label{sec:future_videoa11y}

\subsubsection{Baseline Models}
\label{sec:baseline-videoa11y}
We assessed four open-source models, including Video-LLaVA-7B~\cite{videollava}, VILA1.5-40B~\cite{vila}, LLaVA-NeXT-Video-32B~\cite{llavanext}, and LLaVA-OneVision-72B~\cite{llavaone}. To ensure a fair comparison, we adhered to the original settings for all models, including the number of frames and inference hyperparameters. We then used each MLLM to generate descriptions for the VideoA11y-40K test set without incorporating VideoA11y.

%We generated video descriptions using each MLLM on the VideoA11y-40K test set without applying VideoA11y. 

%Video descriptions were then generated for the VideoA11y-40K test set using each MLLM without incorporating VideoA11y.

%These descriptions were then used to evaluate VideoA11y  (Section~\ref{sec:future_videoa11y}), and the fine-tuned models' performance on VideoA11y-40K (Section~\ref{sec:benchmarking}).
%We then used each MLLM to generate descriptions for videos in the VideoA11y-40K test set with and without VideA11y for the experiments both for VideoA11y and VideoA11y-40K.

\subsubsection{Results on Standard Metrics}
\label{sec:future_nlp_metrics}
Table~\ref{tab:future_nlp_metrics} presents the results based on standard NLP metrics. Notably, all models show improvements across all metrics after integrating VideoA11y, with SPICE showing the largest increase. This suggests that VideoA11y enables models to generate descriptions that are semantically closer to the ground truth, even when the exact wording differs.

\begin{table*}[ht]
\centering
\footnotesize
\caption{Comparison of standard NLP metrics for different models with and without VideoA11y on a held-out test set. \textbf{Bold} numbers indicate better performance for each model. +VA: w/  VideoA11y, -VA: w/o VideoA11y.}
\label{tab:future_nlp_metrics}
\begin{tabular}{l|c|cc cc cc cc cc cc}
\toprule
 &  & \multicolumn{2}{c}{Bleu\_1} & \multicolumn{2}{c}{Bleu\_4} & \multicolumn{2}{c}{METEOR} & \multicolumn{2}{c}{ROUGE\_L} & \multicolumn{2}{c}{CIDEr} & \multicolumn{2}{c}{SPICE} \\
Model & Frames & -VA & +VA & -VA & +VA & -VA & +VA & -VA & +VA & -VA & +VA & -VA & +VA \\
\midrule
Video-LLaVA-7B       & 8  & 12.28 & \textbf{12.73} & 1.08 & \textbf{1.55} & 12.99 & \textbf{13.78} & 16.06 & \textbf{17.27} & 0.63 & \textbf{2.44} & 11.09 & \textbf{14.39}\\
VILA1.5-40B          & 8  & 18.96 & \textbf{20.81} & 4.08 & \textbf{4.57} & 12.55 & \textbf{13.39} & 20.15 & \textbf{21.68} & 8.87 & \textbf{10.97} & 17.72 & \textbf{20.80}\\
LLaVA-NeXT-Video-32B & 32 & 22.57 & \textbf{24.48} & 4.93 & \textbf{5.34} & 20.59 & \textbf{21.99} & 22.54 & \textbf{23.50} & 3.06 & \textbf{3.21} & 19.49 & \textbf{22.14}\\
LLaVA-OneVision-72B  & 24 & 21.01 & \textbf{32.25} & 2.93 & \textbf{7.01} & 15.99 & \textbf{17.73} & 18.75 & \textbf{23.50} & 1.55 & \textbf{14.59} & 13.90 & \textbf{21.32}\\
\bottomrule
\end{tabular}
\end{table*}

\begin{table*}[h]
\centering
\footnotesize
\caption{Comparison of custom metrics for different models with and without VideoA11y on a held-out test set. \textbf{Bold} numbers indicate better performance for each model. +VA: w/ VideoA11y, -VA: w/o VideoA11y.}
\label{tab:future_custom_metrics}
\begin{tabular}{l|c|cccccccc}
\toprule
& & \multicolumn{2}{c}{Descriptive} & \multicolumn{2}{c}{Objective} & \multicolumn{2}{c}{Accurate} & \multicolumn{2}{c}{Clear} \\
Model & Frames & -VA & +VA & -VA & +VA & -VA & +VA & -VA & +VA \\
\midrule
Video-LLaVA-7B       & 8  & 2.72 & \textbf{2.89} & 2.49 & \textbf{2.64} & 1.70 & \textbf{2.10} & 2.70 & \textbf{2.91}\\
VILA1.5-40B          & 8  & 2.35 & \textbf{2.38} & 3.21 & \textbf{3.48} & 2.45 & \textbf{2.52} & 2.87 & \textbf{3.02}\\
LLaVA-OneVision-72B  & 24 & 3.07 & \textbf{3.18} & 3.18 & \textbf{3.46} & 2.32 & \textbf{2.70} & 3.17 & \textbf{3.49}\\
LLaVA-NeXT-Video-32B & 32 & 3.68 & \textbf{3.91} & 3.34 & \textbf{3.39} & 2.76 & \textbf{2.94} & 3.67 & \textbf{3.95}\\
\bottomrule
\end{tabular}
\end{table*}

\begin{table*}[h]
\centering
\footnotesize
\caption{Comparison of standard NLP metrics for different models on a held-out test set. \textbf{Bold} number indicate the best performance, and \underline{underlined} number indicate the second best performance.}
\label{tab:nlp_metrics}
\begin{tabular}{l|c|cccccccc}
\toprule
Model & Frames & Bleu\_1 & Bleu\_4 & METEOR & ROUGE\_L & CIDEr & SPICE \\
\midrule
Video-LLaVA-7B & 8 & 12.73 & 1.55 & 13.78 & 17.27 & 2.44 & 14.39 \\ 
VILA1.5-40B & 8 & 20.81 & 4.57 & 13.39 & 21.68 & 10.97 & 20.80 \\ 
LLaVA-NeXT-Video-32B & 32 & 24.48 & 5.34 & \underline{21.99} & 23.50 & 3.21 & 22.14 \\ 
LLaVA-OneVision-72B & 24 & 32.25 & 7.01 & 17.73 & 23.50 & 14.59 & 21.32 \\ 
\midrule
VideoA11y-7B & 8 & \underline{39.95} & \underline{12.46} & 21.11 & \underline{29.90} & \underline{35.82} & \underline{26.98} \\ 
VideoA11y-32B & 32 & \textbf{41.87} & \textbf{13.95} & \textbf{22.42} & \textbf{31.46} & \textbf{40.29} & \textbf{29.20} \\ 
\bottomrule
\end{tabular}
\end{table*}

\begin{table*}[ht]
\centering
\footnotesize
\caption{Comparison of custom metrics for different models on a held-out test set. \textbf{Bold} number indicate the best performance, and \underline{underlined} number indicate the second best performance.}
\label{tab:custom_metrics}
\begin{tabular}{l|c|cccc}
\toprule
Model & Frames & Descriptive & Objective & Accurate & Clear\\
\midrule
Video-LLaVA-7B       & 8 & 2.89 & 2.64 & 2.10 & 2.91 \\
VILA1.5-40B          & 8 & 2.38 & 3.48 & 2.52 & 3.02 \\
LLaVA-OneVision-72B  & 24 & 3.18 & 3.46 & 2.70 & 3.49 \\
LLaVA-NeXT-Video-32B & 32 & \underline{3.91} & 3.39 & 2.94 & \underline{3.95} \\
\midrule
VideoA11y-7B & 8 & 3.45 & \underline{3.72} & \underline{2.98} & 3.82 \\
VideoA11y-32B & 32 & \textbf{3.98} & \textbf{3.94} & \textbf{3.06} & \textbf{3.97} \\
\bottomrule
\end{tabular}
\end{table*}

%Specifically, we performed experiments on the four open-sourced MLLMs. Using the same held-out test set from Section~\ref{sec:benchmarking}, we assessed model performance with and without VideoA11y integration. 

\subsubsection{Results on Custom Metrics}
\label{sec:future_custom_metrics}
We followed the method outlined in recent work~\citep{videollava, mmbench} to use GPT-4o as an evaluator for video descriptions. In this evaluation framework, GPT-4o treats VideoA11y-40K descriptions as the ground truth and assesses descriptions generated by other models on the four accessibility metrics, %introduced in Section~\ref{sec:metrics}—descriptiveness, objectivity, accuracy, and clarity—
each rated on a scale from 1 to 5. %Table~\ref{tab:future_custom_metrics} reports the performance of these models on custom metrics tailored for accessibility. 
Table~\ref{tab:future_custom_metrics} results indicate consistent improvements across all metrics and all evaluated models when applying VideoA11y. Overall, VideoA11y yields larger improvements in the accurate and clear metrics. Enhanced accuracy ensures that descriptions faithfully capture the video content without including misleading or irrelevant information, while improved clarity provides well-structured and easily comprehensible descriptions. These improvements enable BLV users to better understand and engage with video content.

%Beyond these metric-based improvements, we observed a noteworthy reduction in repetition errors. Repetition, defined as the model repeatedly outputting the same sentence until reaching the maximum token limit, was significantly mitigated with VideoA11y. For example, the number of repetitions for Video-LLaVA-7B decreased from 470 to 276, and for VILA1.5-40B, from 48 to only 4, after applying VideoA11y. These results further emphasize the robustness of VideoA11y in enhancing both quantitative performance and qualitative behavior across a range of MLLMs.

%This validates that VideoA11y can be widely applied to any MLLM, ensuring its continued relevance and robustness as technology evolves.

\subsection{Benchmarking VideoA11y-40K}
\label{sec:benchmarking}

\subsubsection{Baseline Models}
\label{sec:baseline-40k}
We evaluated the same four open-source models described in Section~\ref{sec:baseline-videoa11y} adhering to their original settings. VideoA11y then used each MLLM to generate descriptions for the VideoA11y-40K test set.

\subsubsection{Fine-tuning Models on VideoA11y-40K}
We fine-tuned Video-LLaVA-7B and LLaVA-NeXT-Video-32B on VideoA11y-40K. We refer to the fine-tuned models as VideoA11y-7B and VideoA11y-32B. 
For training, we employed Lora fine-tuning with a configuration of rank 128 and alpha 256. Our fine-tuning parameters included 10 epochs, a learning rate of 2e-5, a batch size of 4 per device, and a maximum model length of 32,768. VideoA11y then used the fine-tuned models to generate descriptions for the videos in the VideoA11y-40K test set.

\subsubsection{Results on Standard Metrics}
\label{sec:nlp_metrics}

We used our dataset to compute standard metrics for descriptions generated by baseline and fine-tuned models. Table~\ref{tab:nlp_metrics} shows that VideoA11y-32B, followed by VideoA11y-7B, significantly outperforms other MLLMs in generating accessible video descriptions in all metrics. These results highlight VideoA11y-32B's ability to generate accurate, detailed, and semantically rich video descriptions, thereby enhancing video accessibility and user engagement.

\subsubsection{Results on Custom Metrics}
\label{sec:custom_metrics}  
Table~\ref{tab:custom_metrics} shows the ratings provided by the GPT-4o evaluator, indicating that VideoA11y-32B achieves the highest scores in all metrics. These results highlight the effectiveness of the VideoA11y-32B model in generating high-quality video descriptions for BLV users and confirm the value of the VideoA11y-40K dataset in improving video accessibility in machine learning models. The GPT-4o evaluator's prompt is shown in Appendix~\ref{app:prompt}.

\section{Discussion}
\label{sec:discussion}
We present the answers to our research questions and discuss the value of VideoA11y and VideoA11y-40K for BLV users and video accessibility research. Moreover, we highlight the limitations of our work and outline future directions.

\subsection{Reflection on Research Questions}
We conducted five studies and four technical experiments to comprehensively assess the effectiveness of VideoA11y and VideoA11y-40K, focusing on answering three research questions: (1) How do VideoA11y descriptions compare in quality to those created by novice and trained human describers? (2) How do professional describers and BLV users evaluate and prefer VideoA11y descriptions compared to human descriptions? (3) Can the VideoA11y-40K dataset enhance SOTA open-source MLLMs to generate high-quality video descriptions specifically tailored for BLV individuals? Our study results consistently demonstrated the value of VideoA11y in addressing these questions. For RQ1, results from Study 2 showed that VideoA11y significantly outperformed novice human annotations in all metrics. In addition, in Study 3 and Study 4 results, VideoA11y descriptions were similar to high-quality descriptions produced by trained humans, as evidenced by evaluations from novice MTurk evaluators and professional describers. For RQ2, results from Study 4 and Study 5 demonstrated that both professional describers and BLV users preferred VideoA11y descriptions over human annotations. In Study 4, professional describers praised VideoA11y’s narrative style and detailed visual representation, highlighting its alignment with professional standards. %Interestingly, some human descriptions were misidentified as AI-generated due to their objectiveness, further underscoring VideoA11y’s ability to match professional expectations. 
%Additionally, results from Study 5 showed that VideoA11y aligned well with the standards of BLV individuals. 
In Study 5, BLV users rated VideoA11y descriptions significantly higher than novice human annotations in all metrics and strongly preferred VideoA11y in over 90\% of cases, enhancing their video comprehension and enjoyment. The BLV users' comments highlighted that VideoA11y descriptions had high clarity, matched with unfolding events, and were more balanced and impartial than human annotations. For RQ3, we compared competitive baseline models with two models fine-tuned on VideoA11y-40K and found that the models fine-tuned on VideoA11y-40K significantly outperformed the baselines in generating accessible video descriptions.
\subsection{Implications of Our Method, Dataset, and Benchmark for Video Accessibility}

VideoA11y enables supporting video accessibility for BLV users at scale. 
While hundreds of video description models have been introduced in the computer vision community over the past decade~\cite{aafaq2019video}, none have addressed the real-world societal application of audio descriptions. This gap emphasized the importance and novelty of VideoA11y in accelerating progress in this underexplored area. Moreover, collecting high-quality human annotations for videos is hard to scale. When we collected high-quality human annotations for 47 videos in Study 3 (Section~\ref{sec:study3}), each annotator took an average of 3-4 minutes to describe 1 minute of a video. Even with meticulous effort, human annotations could not surpass VideoA11y's descriptions. %sometimes unintentionally introduced typos or grammatical errors in the video descriptions. These oversights were noted in Study 4 (Section~\ref{sec:studypd}), where and professional audio describers highlighted their significant impact on BLV users' experience. 
In contrast, VideoA11y accelerates the creation of video descriptions, ensuring error-free outputs without grammatical mistakes, and is applicable to any video content, as demonstrated by our studies involving over 197 videos from 15 different categories. It can also be extended to incorporate additional guidelines and MLLMs that handle other modalities (e.g., audio). Thus, VideoA11y has the potential to be used on online video-sharing platforms (e.g., YouTube and TikTok) to create accessible content for BLV users.

VideoA11y produced minimal hallucinations. In the context of video description, hallucination refers to inaccuracies where the description includes details not present in the actual video content~\cite{li2024vidhalluc}. When VideoA11y was applied without human annotations as a reference (VideoA11y w/o HA), there were occasional instances where the model introduced actions or details not found in the video (see examples in Appendix~\ref{app:qualitative}). However, when human annotations were incorporated as references, VideoA11y showed a reduction in hallucinations, as evidenced by the `accuracy' ratings of over 4.2 out of 5 by 300 sighted users in Study 1 and Study 2. Additionally, we observed that BLV users had heightened sensitivity to the audio components of the videos, enabling them to detect inconsistencies between the audio content and the provided descriptions. Therefore, the high `accuracy' scores by BLV users further demonstrate that VideoA11y descriptions are highly accurate.

VideoA11y can also shift the role of audio describers from generating descriptions to providing guidance and verification for MLLM-generated content. The effectiveness of VideoA11y reflects the richness and value of AD best practices and guidelines developed by professional describers over the past decades. In fact, several professional audio describers emphasized the importance of our four custom metrics—descriptive, objective, accurate, and clear—and noted that all are essential for evaluating and producing high-quality video descriptions. While larger datasets can improve MLLM output, our studies demonstrated that providing effective guidance to the model was equally important. Our results suggest a pathway for integrating MLLMs with the AD professional community to develop a human-AI collaboration~\cite{hitl_blv_2020} workforce for accessibility. For instance, in our experiments in Section~\ref{sec:future_videoa11y}, we demonstrated that applying the existing 42 guidelines improved MLLM performance across all metrics without any additional training. Building on this success, audio describers could further enhance the list of guidelines by adding new and revised rules for MLLMs, creating specific guidelines for different video genres (e.g., entertainment vs. people and vlogs, long- vs. short-form videos), and verifying the model's output. The model could also provide the list of guidelines used to generate a description, further facilitating the human verification and guidance process.

The VideoA11y-40K dataset, along with its benchmark, can support the development of future computer vision models. As the largest dataset on video accessibility, VideoA11y-40K enables researchers to train models across various video genres, and our benchmark can assess the robustness of these models. Our experiments in Section~\ref{sec:experiments} demonstrated that fine-tuning open-source MLLMs on VideoA11y-40K led to significant improvements compared to baseline models in both standard and custom metrics. While the performance of the fine-tuned models does not yet match that of the GPT-4V, they provide a more cost-effective and scalable solution. The scalability of fine-tuned open-source models is key to widespread adoption, enabling the continuous creation of tailored descriptions for BLV users at a significantly lower cost.

\subsection{Limitations and Future Work}
\label{sec:limitations}
When human annotations are absent as references, relying solely on AD guidelines for VideoA11y to generate descriptions can lead to hallucinations. In our tests, these inaccuracies were mostly related to minor scene details. However, the broader impact of such inaccuracies in video descriptions on a larger scale remains largely unexplored. For example, during a public health crisis, inaccurate descriptions could prevent BLV individuals from receiving crucial information needed to make informed decisions. Additionally, VideoA11y is vulnerable to injection attacks~\cite{liu2023prompt}, where human annotations might include false or harmful content, potentially spreading misinformation or negatively affecting the well-being of BLV users. Future research could explore the impact of these hallucinations across different content types and investigate ways to reduce inaccuracies. This may include applying direct preference optimization (DPO) during training to minimize hallucinations~\cite{dpo}, using helper models for extracting specific information (e.g., object recognition~\cite{ssnn}, optical character recognition~\cite{dtrocr}) as input for prompts, or incorporating feedback from sighted volunteers to enhance the reliability of AI-generated descriptions.

Another limitation of VideoA11y is the lack of customization options tailored to the specific preferences and needs of BLV individuals. While our approach relies on general AD guidelines, it does not currently support personalized adjustments based on individual user preferences~\citep{tentative_criteria, context_dependent, ad_customization}, such as preferred levels of detail or focus on specific types of content (e.g., action vs. emotional tone). Future research could gather more data on BLV users' preferences, such as the desired level of detail, focus of descriptions, and preferred style or tone across various video categories. These insights could enable VideoA11y to dynamically adapt to individual preferences, further enhancing comprehension and enjoyment for BLV users.
 
Lastly, VideoA11y currently lacks the ability to control the length and timing of descriptions to fit seamlessly within the natural pauses of a video, which is crucial for supporting inline descriptions. BLV users generally prefer inline descriptions that are integrated into the video flow without pausing or interrupting the experience~\cite{context_dependent}. To enable this, the system would need to dynamically adjust the length of descriptions and identify appropriate moments in the video to present them~\cite{rescribe}. While these aspects were not the focus of our current work, future studies could integrate VideoA11y with systems like Rescribe~\cite{rescribe} to support inline descriptions, enhancing the flexibility and usability of VideoA11y for BLV users and its integration into video-watching platforms.

\section{Conclusion}
\label{sec:conclusion}
This paper addresses a critical gap in video understanding research by developing a novel approach to enhance video accessibility for BLV individuals. We introduced VideoA11y, a method that leverages multimodal large language models and accessibility guidelines to generate video descriptions specifically tailored to the unique needs of BLV users. Rigorous experiments showed that VideoA11y outperformed both novice and trained human describers in terms of clarity, accuracy, objectivity, and descriptiveness, achieving higher satisfaction among BLV users. In addition, we presented the VideoA11y-40K dataset, the largest and most comprehensive video description dataset, comprising 40,000 videos described according to AD guidelines. Benchmarking VideoA11y-40K using both standard and custom metrics demonstrates that fine-tuned MLLMs on this dataset generate more accessible descriptions for BLV users, underscoring the potential of MLLMs to scale video content accessibility.

\begin{acks}
This research was supported by the National Eye Institute (NEI) of the National Institutes of Health (NIH) under award number R01EY034562. The content is solely the responsibility of the authors and does not necessarily represent the official views of the NIH. 
\end{acks}

\bibliographystyle{ACM-Reference-Format}
\bibliography{reference}

\newpage
\appendix
%TC:ignore

\section{Audio Descriptions Guidelines}
\label{app:ad}
The list below shows the complete 42 audio description (AD) guidelines we curated for VideoA11y.

\setlength{\fboxsep}{8pt} 
\setlength{\fboxrule}{0.5mm}
\vspace{0.2cm}
\noindent
\fcolorbox{black}{lightgray}{ 
    \begin{minipage}{0.94\linewidth}
    \raggedright
    Instruction \#1. Avoid over-describing — Do not include non-essential visual details.
    
    Instruction \#2. Description should not be opinionated unless content demands it.
    
    Instruction \#3. Choose level of detail based on plot relevance when describing scenes.
    
    Instruction \#4. Description should be informative and conversational, in present tense and third-person omniscient.
    
    Instruction \#5. The vocabulary should reflect the predominant language/accent of the program and should be consistent with the genre and tone of the content while also mindful of the target audience. Vocabulary used should ensure accuracy, clarity, and conciseness.

    Instruction \#6. Consider historical context and avoid words with negative connotations or bias.

    Instruction \#7. Pay attention to verbs — Choose vivid verbs over bland ones with adverbs.

    Instruction \#8. Use pronouns only when clear whom they refer to.

    Instruction \#9. Use comparisons for shapes and sizes with familiar and globally relevant objects.

    Instruction \#10. Maintain consistency in word choice, character qualities, and visual elements for all audio descriptions.

    Instruction \#11. Tone and vocabulary should match the target audience's age range.

    Instruction \#12. Ensure no errors in word selection, pronunciation, diction, or enunciation.

    Instruction \#13. Start with general context, then add details.

    Instruction \#14. Describe shape, size, texture, or color as appropriate to the content.

    Instruction \#15. Use first-person narrative for engagement if required to engage the audience.

    Instruction \#16. Use articles appropriately to introduce or refer to subjects.

    Instruction \#17. Prefer formal speech over colloquialisms, except where appropriate.

    Instruction \#18. When introducing new terms, objects, or actions, label them first, and then follow with the definitions.

    Instruction \#19. Describe objectively without personal interpretation or comment. Also, do not censor content.

    Instruction \#20. Deliver narration steadily and impersonally (but not monotonously), matching the program's tone.

    Instruction \#21. It can be important to add emotion, excitement, lightness of touch at different points. Adjust style for emotion and mood according to the program's genre.

    Instruction \#22. If it is children’s content, tailor language and pace for children's TV, considering audience feedback. 

    Instruction \#23. Do not alter, filter, or exclude content. You should describe what you see. Try to seek simplicity and succinctness in your description.
    \end{minipage}
}

\setlength{\fboxsep}{8pt} 
\setlength{\fboxrule}{0.5mm}
\noindent
\fcolorbox{black}{lightgray}{ 
    \begin{minipage}{0.94\linewidth}
    \raggedright
    Instruction \#24. Prioritize what is relevant when describing action as to not affect user experience.
    
    Instruction \#25. Include location, time, and weather conditions when relevant to the scene or plot.
    
    Instruction \#26. Focus on key content for learning and enjoyment when creating audio descriptions. This is so that the intention of the program is conveyed.

    Instruction \#27. When describing an instructional video/content, describe the sequence of activities first.
    
    Instruction \#28. For a dramatic production, include elements such as style, setting, focus, period, dress, facial features, objects, and aesthetics.

    Instruction \#29. Describe what is most essential for the viewer to know in order to follow, understand, and appreciate the intended learning outcomes of the video/content.

    Instruction \#30. The description should describe characters, locations, on-screen action, and on-screen information.

    Instruction \#31. Describe only what a sighted viewer can see.

    Instruction \#32. Describe main and key supporting characters' visual aspects relevant to identity and personality. Prioritize factual descriptions of traits like hair, skin, eyes, build, height, age, and visible disabilities. Ensure consistency and avoid singling out characters for specific traits. Use person-first language.

    Instruction \#33. If unable to confirm or if not established in the plot, do not guess or assume racial, ethnic or gender identity.

    Instruction \#34. When naming characters for the first time, aim to include a descriptor before the name (e.g., "a bearded man, Jack").

    Instruction \#35. Description should convey facial expressions, body language and reactions.

    Instruction \#36. When important to the meaning / intent of content, describe race using currently-accepted terminology.

    Instruction \#37. Avoid identifying characters solely by gender expression unless it offers unique insights not apparent otherwise to low vision viewers.

    Instruction \#38. Describe character clothing if it enhances characterization, plot, setting, or genre enjoyment.

    Instruction \#39. If text on the screen is central to understanding, establish a pattern of on-screen words being read. This may include making an announcement, such as "Words appear".

    Instruction \#40. In the case of subtitles, the describer should read the translation after stating that a subtitle appears.

    Instruction \#41. When shot changes are critical to the understanding of the scene, indicate them by describing where the action is or where characters are present in the new shot.

    Instruction \#42. Provide description before the content rather than after.
    \end{minipage}
}

\section{Prompts and Implementation Details}
\label{app:prompt}
\subsection{Prompt for GPT-4V}
The following prompt was employed for the \textit{GPT-4V} method in Study 2, as detailed in Section~\ref{sec:study2}.

\setlength{\fboxsep}{8pt} 
\setlength{\fboxrule}{0.5mm}
\vspace{0.2cm}
\noindent
\fcolorbox{black}{lightgray}{ 
    \begin{minipage}{0.94\linewidth}
    \raggedright
    Imagine your role is to generate descriptions for videos. You will watch a sequence of keyframes 
    from a video and craft a description based on these keyframes.
    \end{minipage}
}

\subsection{Prompt for GPT-4V w/ HA}
The following prompt was employed for the \textit{GPT-4V w/ HA} method in Study 2, as detailed in Section~\ref{sec:study2}.

\setlength{\fboxsep}{8pt} 
\setlength{\fboxrule}{0.5mm}
\vspace{0.2cm}
\noindent
\fcolorbox{black}{lightgray}{ 
    \begin{minipage}{0.94\linewidth}
    \raggedright
    Imagine your role is to generate descriptions for videos. You will watch a sequence of keyframes 
    from a video and read the current description of this video. Your task is to revise the description.
    \end{minipage}
}

\subsection{Prompt for VideoA11y w/o HA}
The following prompt was employed for the \textit{VideoA11y (LLaVA) w/o  HA} and the \textit{VideoA11y (GPT) w/o HA} method in Study 1 (as outlined in Section~\ref{sec:study1}), and \textit{VideoA11y w/o HA} method in Study 2 (Section~\ref{sec:study2}).

\setlength{\fboxsep}{8pt} 
\setlength{\fboxrule}{0.5mm}
\vspace{0.2cm}
\noindent
\fcolorbox{black}{lightgray}{ 
    \begin{minipage}{0.94\linewidth}
    \raggedright
    Imagine your role is to generate descriptions for videos to make them accessible to blind and 
    low vision individuals. You will watch a sequence of keyframes from a video. Based on these 
    keyframes, craft a description. You must follow all the given instructions. You should avoid any 
    prefatory language, such as `the video shows'. Output your result as a dictionary format: 
    \{``Video\_Category'': A string representing the category of video you believe it to be, 
    ``Revised\_Desc'': A string of description.\}
    \end{minipage}
}

\subsection{Prompt for VideoA11y}
The following prompt was employed for the \textit{VideoA11y (LLaVA)} and the \textit{VideoA11y (GPT)} methods in Study 1 (as outlined in Section~\ref{sec:study1}), and the VideoA11y method in Study 2 (Section~\ref{sec:study2}).

\setlength{\fboxsep}{8pt} 
\setlength{\fboxrule}{0.5mm}
\vspace{0.2cm}
\noindent
\fcolorbox{black}{lightgray}{ 
    \begin{minipage}{0.94\linewidth}
    \raggedright
    Imagine your role is to generate descriptions for videos to make them accessible to blind and low vision individuals. You will watch a sequence of keyframes from a video and read the current description of this video. Your task is to revise the current description. You must follow all the given instructions. Output your result in a dictionary format: \{``Video\_Category'': A string representing the category of video you believe it to be, ``Revised\_Desc'': A string of revised description.\}
    \end{minipage}
}

\subsection{Prompt for GPT-4o Evaluator}
The following prompt was employed for the GPT-4o evaluator used in technical experiments (as outlined in Section~\ref{sec:custom_metrics}).

\setlength{\fboxsep}{8pt} 
\setlength{\fboxrule}{0.5mm}
\vspace{0.2cm}
\noindent
\fcolorbox{black}{lightgray}{ 
    \begin{minipage}{0.94\linewidth}
    \raggedright
    You are an expert evaluator with a deep understanding of video description quality,
    particularly for making content accessible to blind and low vision (BLV) individuals. Your role is to assess and rate video descriptions based on specific metrics.
    \end{minipage}
}

\setlength{\fboxsep}{8pt} 
\setlength{\fboxrule}{0.5mm}
\vspace{0.2cm}
\noindent
\fcolorbox{black}{lightgray}{ 
    \begin{minipage}{0.94\linewidth}
    \raggedright
    Task:

    I have two video descriptions: one is the ground truth, and the other is generated by a model. Additionally, I have four evaluation metrics: Descriptive, Objective, Accurate, and Clear. Please evaluate the model-generated description using the above metrics. Provide a rating from 1 to 5 for each metric, and briefly explain the reasons for each rating.
    
    ~\\
    Metrics Definition:

    1. Descriptive: A descriptive description should provide vivid details about objects, people, and settings while maintaining a concise narrative flow. It should include essential information about the appearance of individuals, such as their clothing, facial expressions, and actions, and visual properties of objects, such as color and shape. For example, "A smiling woman, wearing a loose white dress, types on a laptop in a softly lit room."

    2. Objective: An objective description should report what is visible without adding personal opinions or assumptions. For instance, describe two people as “a woman and a man” without adding any relationship context unless it is mentioned. It should also avoid guessing personal attributes like racial or gender identities unless explicitly clear.

    3. Accurate: An accurate description should aim for precision in describing visible elements without imagination. It should ensure that all details, such as colors and spatial arrangements, are reported correctly. For instance, "Blue sky with white clouds" instead of "White sky with blue clouds" if that is what appears. Additionally, it should avoid adding unnecessary details that do not contribute to a deeper understanding of the scene.
    
    4. Clear: A clear description should present information in the videos in a way that is easy to follow for blind and low vision individuals. It should describe the object or character’s properties before the actions or relationships with other objects or characters. For example, "A man wearing sunglasses plays the piano." Pronouns should only be used when it is clear who they refer to, and the description should include any on-screen text if it is central to understanding. For instance, "A man in a black polo shirt is speaking. He is in a studio setting with a red couch and a blue background featuring the text ’Talk of the Town’".
    
    ~\\
    Input:
    
    - Ground truth video description: \{desc\_gt\}

    - Model-generated video description: \{desc\_can\}

    ~\\
    Output Format:

    Return the result as a string format dictionary with the following structure:
    
    \{``Descriptive'': [Rating out of 5],
    
    ``Objective'': [Rating out of 5],

    ``Accurate'': [Rating out of 5],
    
    ``Clear'': [Rating out of 5],
    
    ``Reason'': "Your brief explanation here"\}
    \end{minipage}
}

\begin{table*}[h]
  \footnotesize
  \caption{Evaluation metrics used in the user studies. Sighted MTurk and BLV participants reviewed these definitions before rating the video descriptions.}
  \label{evalution-metrics}
  \centering
  \begin{tabular}{lp{0.8\textwidth}}
    \toprule
    \textbf{Metric}     & \textbf{Description} \\
    \toprule
    Descriptive & A descriptive description should provide vivid details about objects, people, and settings while maintaining a concise narrative flow. It should include essential information about the appearance of individuals, such as their clothing, facial expressions, and actions, and visual properties of objects, such as color and shape. For example, "A smiling woman, wearing a loose white dress, types on a laptop in a softly lit room." \\
    \midrule
    Objective  & An objective description should report what is visible without adding personal opinions or assumptions. For instance, describe two people as “a woman and a man” without adding any relationship context unless it is mentioned. It should also avoid guessing personal attributes like racial or gender identities unless explicitly clear. \\ 
    \midrule
    Accurate     & An accurate description should aim for precision in describing visible elements without imagination. It should ensure that all details, such as colors and spatial arrangements, are reported correctly. For instance, "Blue sky with white clouds" instead of "White sky with blue clouds" if that is what appears. Additionally, it should avoid adding unnecessary details that do not contribute to a deeper understanding of the scene.       \\
    \midrule
    Clear     & A clear description should present information in the videos in a way that is easy to follow for blind and low vision individuals. It should describe the object or character’s properties before the actions or relationships with other objects or characters. For example, "A man wearing sunglasses plays the piano." Pronouns should only be used when it is clear who they refer to, and the description should include any on-screen text if it is central to understanding. For instance, "A man in a black polo shirt is speaking. He is in a studio setting with a red couch and a blue background featuring the text 'Talk of the Town'". \\
    \bottomrule
  \end{tabular}
\end{table*}

\section{Metrics Definition}
\label{app:metrics}
Table~\ref{evalution-metrics} provides a comprehensive definition of our four customized metrics for Study 1 (Section~\ref{sec:study1}), Study 2 (Section~\ref{sec:study2}), Study 3 (Section~\ref{sec:study3}), Study 4 (Section~\ref{sec:studypd}), and Study 5 (Section~\ref{sec:studyblv}). These definitions were also presented to participants in all studies (See Appendix~\ref{app:interface}).

\section{User Study Interfaces}
\label{app:interface}

\subsection{User Interface of Studies 1, 2 and 3}
\begin{figure*}
    \centering
    \includegraphics[width=0.94\linewidth]{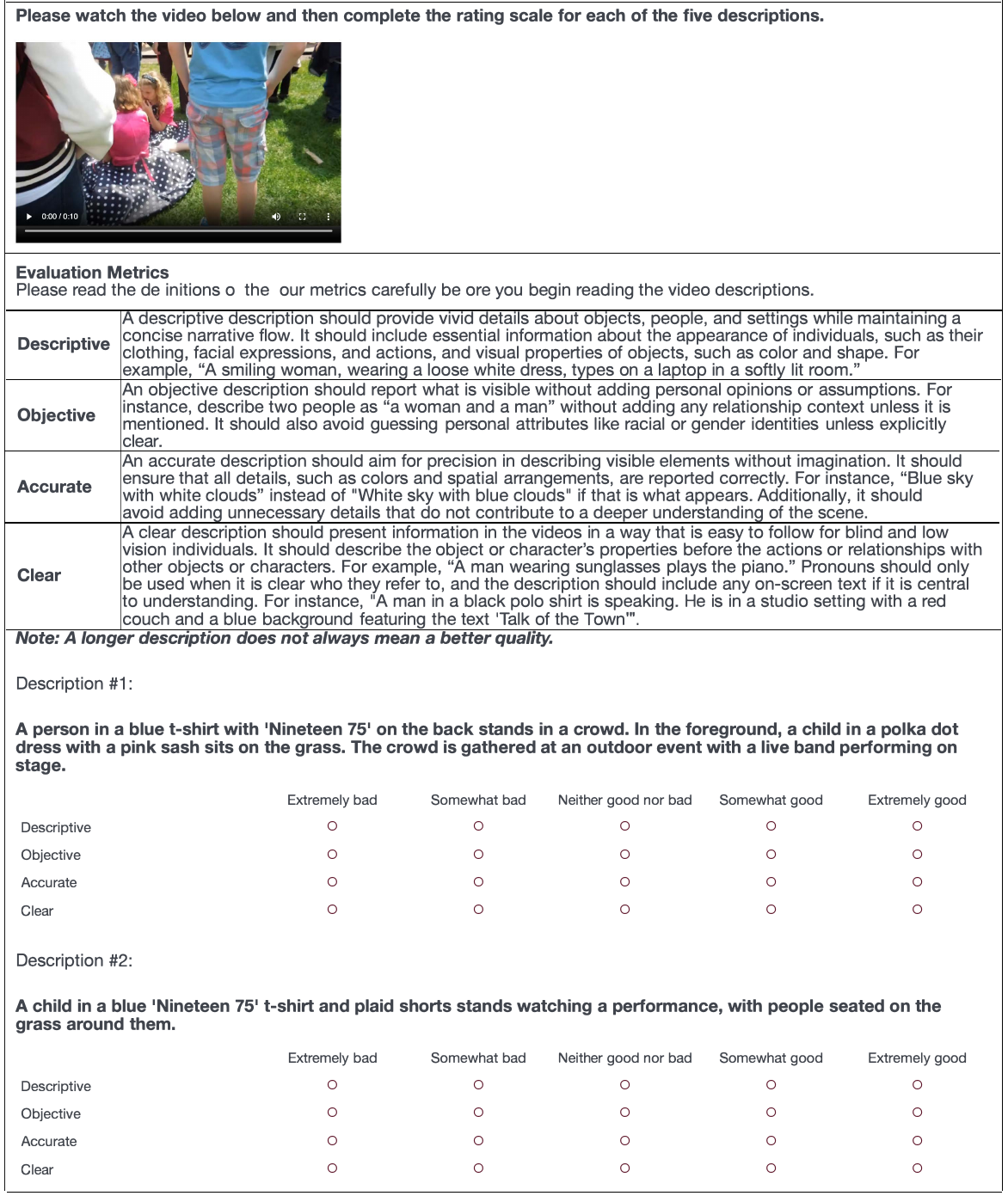}
    \caption{User interface in Studies 1, 2 and 3. MTurk participants in Study 1 rated four video descriptions from different methods, MTurk participants in Study 2 rated five video descriptions generated by different methods. MTurk participants in Study 3 rated two video descriptions generated by VideoA11y and human. Video descriptions were presented to participants in random order.}
    \label{fig:inter_studys}
    \Description{The user interface begins by displaying a video at the top of the page, instructing participants to watch it before proceeding. Below the video, participants are presented with the definitions of four evaluation metrics: Descriptive, Objective, Accurate, and Clear. Participants are asked to carefully read these definitions to understand how each metric should be applied during the evaluation. After familiarizing themselves with the evaluation metrics, participants are shown descriptions of the video content. For each description, they are required to rate it on the four metrics using a Likert scale, ranging from "Extremely bad" to "Extremely good." The evaluation is completed by selecting one rating per metric for each description.}
\end{figure*}

Figure~\ref{fig:inter_studys} illustrates the user interface used in Studies 1, 2 and 3. After providing informed consent on the first page of the online survey, participants watched a video, followed by reading the definitions of the four metrics proposed in Section~\ref{sec:metrics}. Subsequently, they were presented with multiple video descriptions generated by different methods (four in Study 1, five in Study 2, and two in Study 3). Each description was rated from ``Extremely bad'' to ``Extremely good'' based on the aforementioned metrics. To ensure fairness, all video descriptions were presented in a randomized order. This same procedure was then repeated with a long video.

\subsection{User Interface of Study 5}
\begin{figure*}[t!]
   \centering
   \begin{subfigure}[t]{0.45\textwidth}
     \centering
     \raisebox{0.8mm}
     {\includegraphics[width=0.9\linewidth]{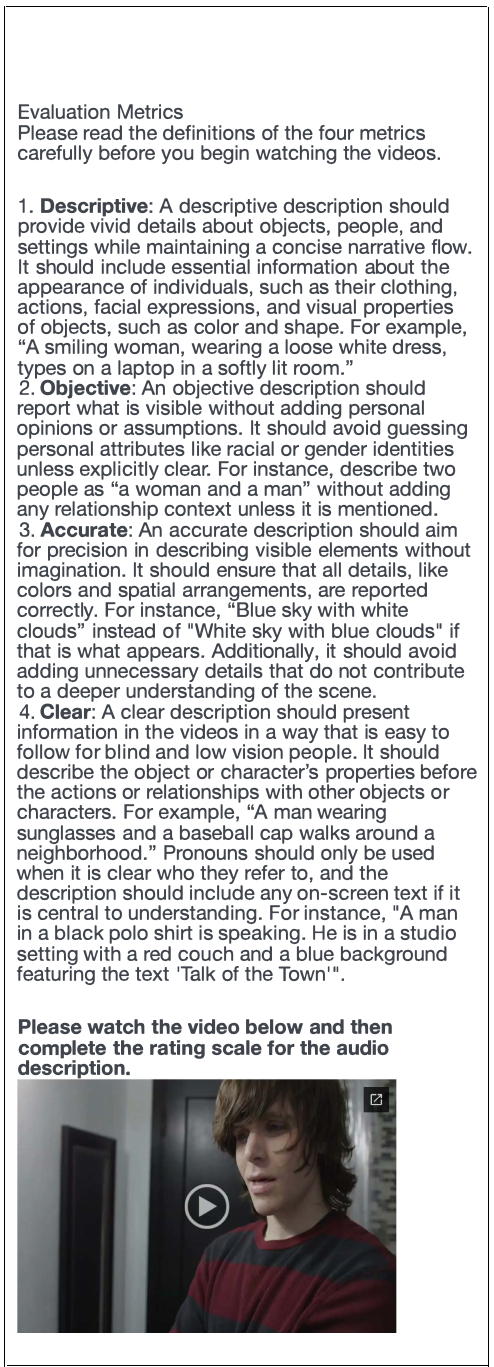}}
   \end{subfigure}\hfill
   \begin{subfigure}[t]{0.45\textwidth}
     \centering
     \includegraphics[width=0.9\linewidth]{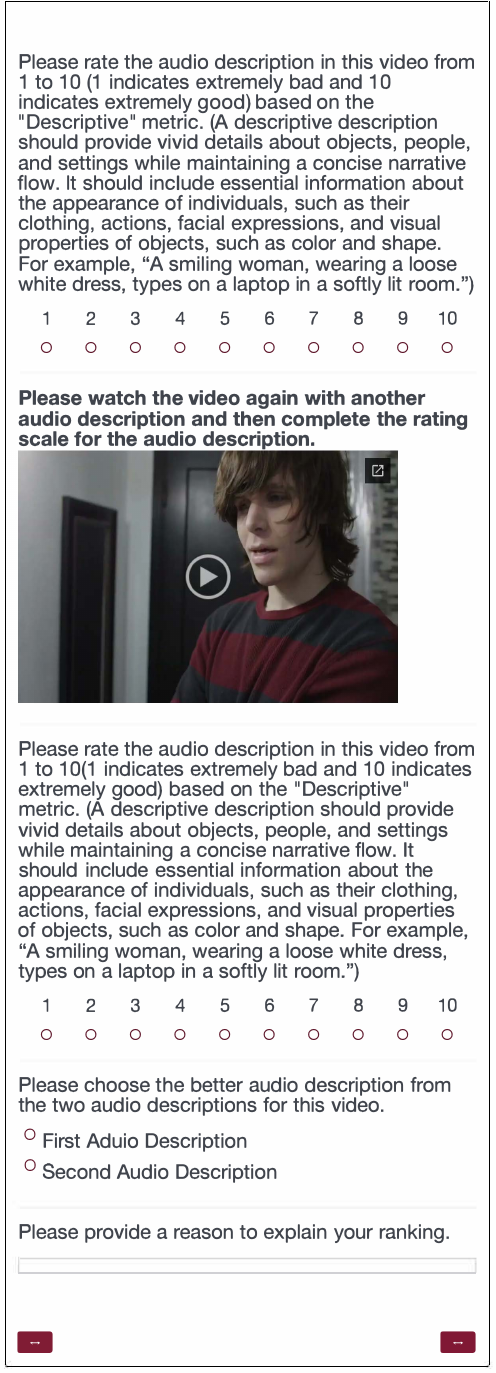}
   \end{subfigure}
  % \vspace{-7mm}
   \caption{User interface in Study 5. We carefully designed all content to be fully accessible to BLV users via screen reader. This figure displays only the rating question for descriptiveness for both video descriptions due to space constraints. In the actual study, participants rated each video description against all four metrics. After evaluating two descriptions for the same video, participants ranked them and justified their rankings.}
   \label{fig:inter_blv}
   \Description{The interface begins with definitions of four evaluation metrics—Descriptive, Objective, Accurate, and Clear—that participants are asked to review. After watching a video with the first audio description, participants rate the audio description on a scale from 1 to 10 for the four metrics. They then watch the video again with a second audio description and rate it similarly. Finally, participants choose the better audio description and provide a reason for their choice.}
\end{figure*}

Figure~\ref{fig:inter_blv} illustrates the user interface used in Study 5. After providing informed consent on the first page of the online survey, BLV participants read the definitions of four metrics proposed in Section~\ref{sec:metrics}. They were then presented with a video paired with a human-annotated video description and they rated the quality of the description on a scale from 1 to 10, for each of the four metrics. This extended rating scale, from 1 to 10, was adopted following feedback from BLV users during pilot testing, who indicated that a 5-point scale was inadequate for BLV individuals to capture the nuanced variations in video descriptions. Subsequently, participants watched the same video accompanied by the VideoA11y-40K description, and again rated it using the same set of metrics. The sequence in which the human-annotated and VideoA11y-40K descriptions were presented was randomized to mitigate order bias. After evaluating both descriptions, participants were asked to rank them and provide justifications for their preferences. This process was replicated across four additional videos to ensure robust assessment.

\subsection{User Interface of Video Category Evaluation Study}
\begin{figure*}[h]
   \centering
   \begin{subfigure}[t]{0.5\textwidth}
     \centering
     \includegraphics[width=1\linewidth]{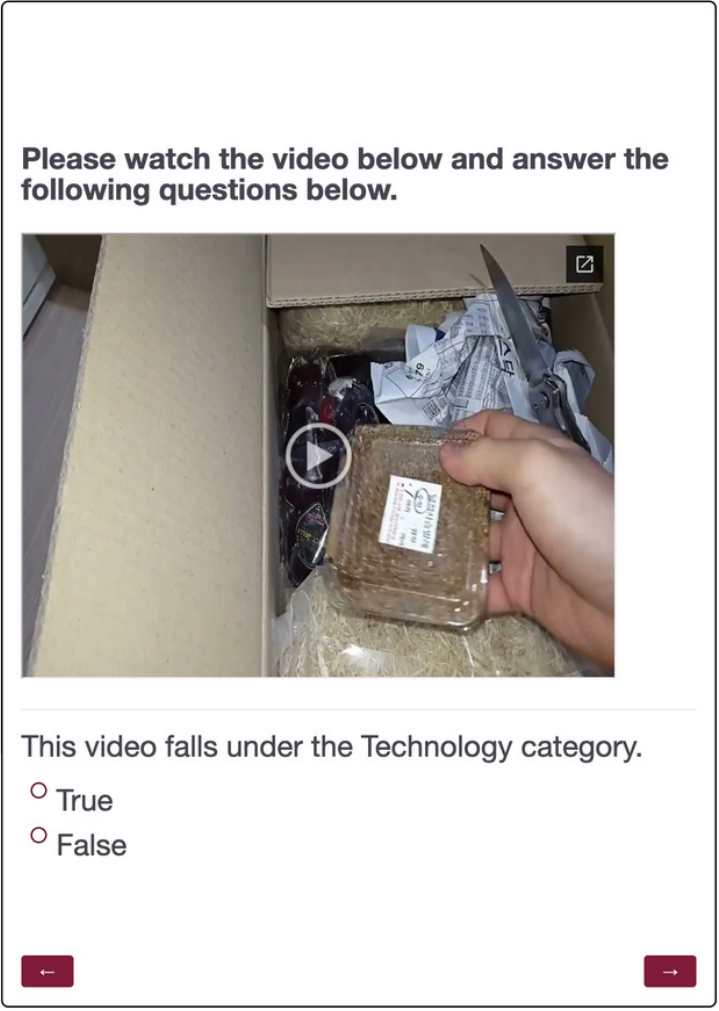} 
     \caption{The interface for collecting responses to determine whether the video category is correct.}
     \label{sub:correct}
   \end{subfigure}\hfill
   \begin{subfigure}[t]{0.491\textwidth}
     \centering
     \raisebox{0.5mm}
     {\includegraphics[width=1\linewidth]{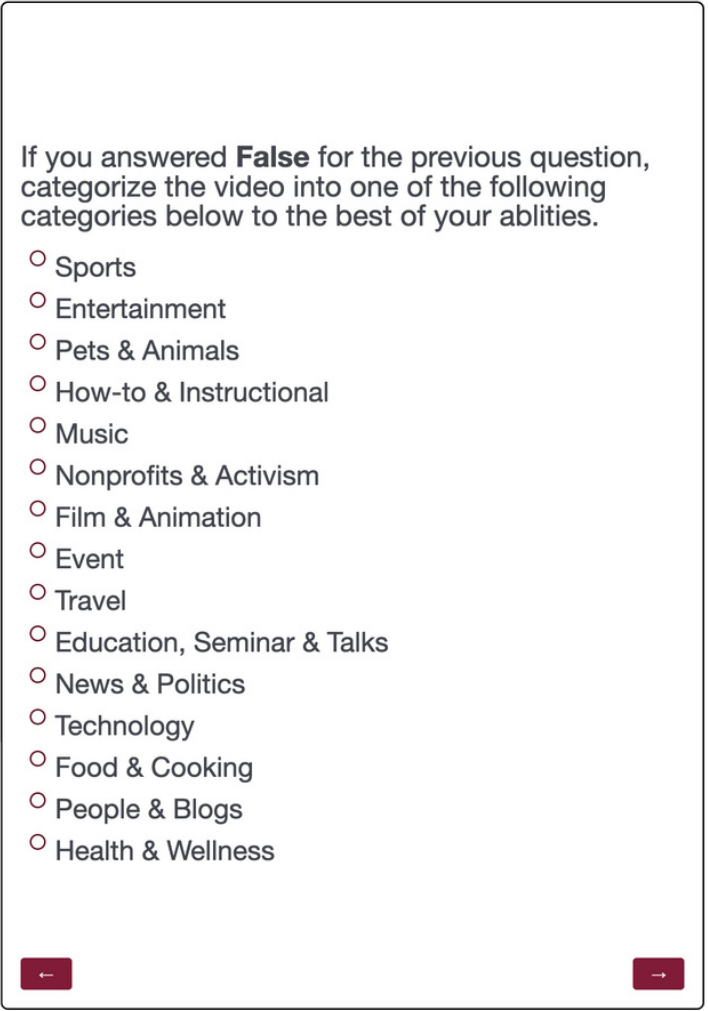}}
     \caption{The interface for collecting responses to re-categorize the video.}
     \label{sub:recategorize}
   \end{subfigure}
   \caption{User interface of the category evaluation study for VideoA11y-40K. (a) MTurk participants watched a video and the category assigned by VideoA11y first and then determined whether the video category was correct. (b) If they selected ``False,'' they were redirected to this page to assign a new category to the video.}
   \label{fig:inter_category}
   \Description{The interface consists of two steps. In the first step (a), participants watch a video and answer whether the assigned category is correct by selecting "True" or "False." If participants select "False," they are directed to the second step (b), where they choose the correct category from a list of 15 options.}
\end{figure*}

Figure~\ref{fig:inter_category} illustrates the user interface for the video category evaluation study. After providing informed consent on the first page of the online survey, participants viewed a video that had been pre-categorized by VideoA11y. They were then tasked with verifying the appropriateness of the assigned category on page~\ref{sub:correct}. If participants deemed the category incorrect (``False''), they were redirected to page~\ref{sub:recategorize}, where they could reassign the video to one of the 15 video categories.

\subsection{User Interface of Demographic Questionnaire}
\begin{figure*}[h]
    \centering
    \includegraphics[width=1\linewidth]{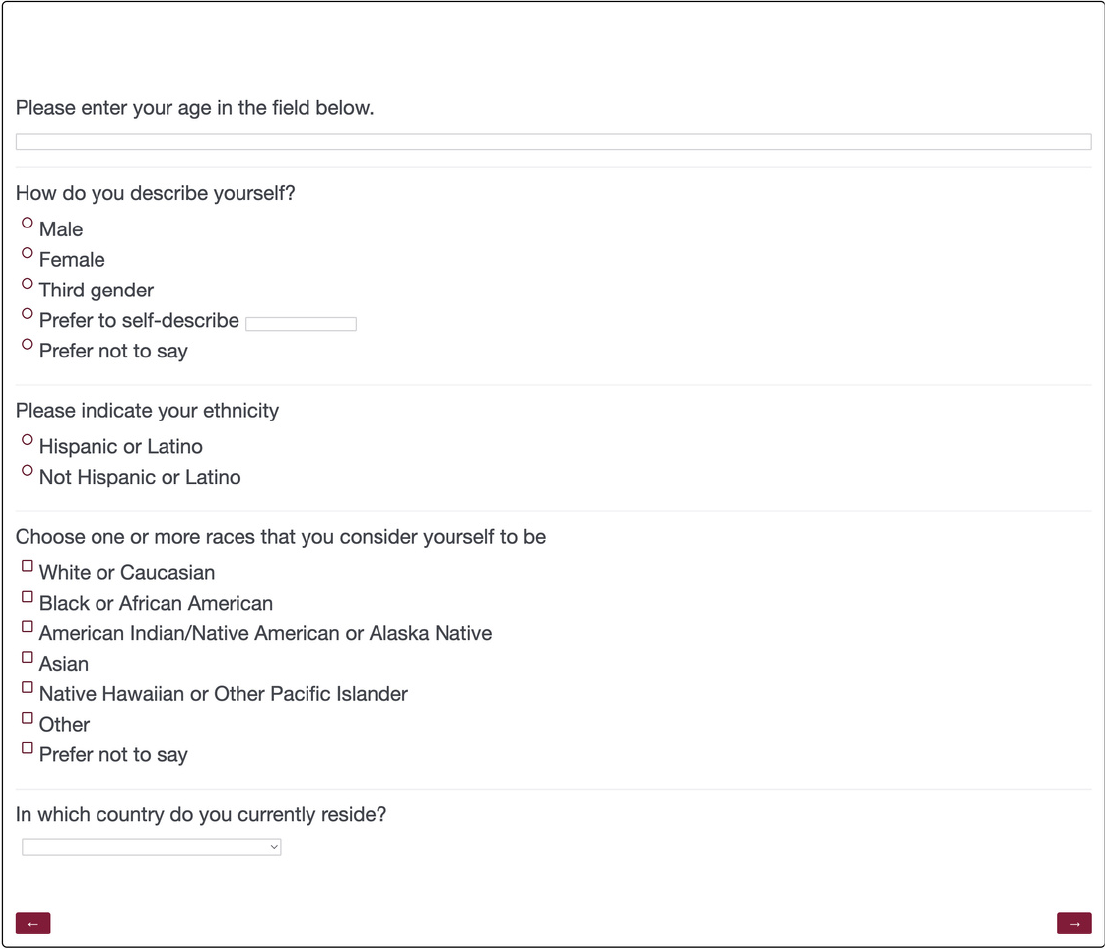}
    \caption{The interface of the demographic questionnaire for all human studies.}
    \label{fig:demographic}
    \Description{The interface displays a demographic questionnaire. Participants are asked to enter their age, followed by questions about gender, ethnicity, race, and country of residence. Options for gender include Male, Female, Third gender, and a self-describe field. Participants are also asked to indicate their ethnicity (Hispanic or Latino, or Not Hispanic or Latino) and select one or more racial categories they identify with, such as White, Black, Asian, or Other. The questionnaire ends with a question about the participant’s country of residence.}
   \end{figure*}

Figure~\ref{fig:demographic} shows the demographic questionnaire used across all studies. The questionnaire was designed to collect data on participants' age, gender, ethnicity, race, and country of residence. Details of the demographic data are presented in the results section of each study in Section~\ref{sec:evaluation}.

\section{Statistical Analysis of the Studies}
\label{app:stat}

\subsection{Study 1: Statistical Analysis}
\label{app:stat_study1}
Table~\ref{tab:pairwise_comparisons_1} provides additional statistical analysis for Study 1 (Section~\ref{sec:study1}). We applied the related-samples Friedman test, followed by post hoc pairwise comparisons, to analyze the data. Statistically significant results ($p<0.05$) are highlighted in bold. These results demonstrate that VideoA11y (GPT) and VideoA11y (GPT) w/o HA show statistically significant improvements over VideoA11y (LLaVA) and VideoA11y (LLaVA) w/o HA in all four metrics.

\subsection{Study 2: Statistical Analysis}
\label{app:stat_study2}
Table~\ref{tab:pairwise_comparisons_2} provides additional statistical analysis for Study 2 (Section~\ref{sec:study2}). The data is analyzed using the related-samples Friedman test, followed by post hoc pairwise comparisons. Significant results ($p<0.05$) are highlighted in bold. The analysis reveals that VideoA11y and VideoA11y w/o HA show statistically significant superiority in all four metrics when compared to Human Annotation, GPT-4V, and GPT-4V w/ HA.

\begin{table*}[t]
\footnotesize
\centering
\caption{Overall pairwise comparisons evaluating VideoA11y on open-source and proprietary MLLMs in Study 1. HA: Human Annotation.}
\label{tab:pairwise_comparisons_1}
\begin{tabular}{>{\raggedright\arraybackslash}p{0.40\linewidth}
>{\raggedright\arraybackslash}p{0.14\linewidth}
>{\raggedright\arraybackslash}p{0.11\linewidth}
>{\raggedright\arraybackslash}p{0.08\linewidth}}
\toprule
\textbf{Condition 1 | Condition 2} & \textbf{Metric} & \textbf{Test Statistics} & \textbf{P Value} \\
\midrule
VideoA11y (GPT) w/o HA | VideoA11y (LLaVA) w/o HA& Descriptive& 4.048  & \textbf{$<$0.001} \\
 & Objective & 5.708 & \textbf{$<$0.001} \\
 & Accurate& 5.313 & \textbf{$<$0.001} \\
 & Clear& 4.095& \textbf{$<$0.001} \\
\midrule
VideoA11y (GPT) w/o HA | VideoA11y (LLaVA)&  Descriptive& 2.546	 & \textbf{0.011} \\
 & Objective & 4.048 & \textbf{$<$0.001} \\
 & Accurate& 4.411 & \textbf{$<$0.001} \\
 & Clear& 3.178 & \textbf{0.001} \\
\midrule
VideoA11y (GPT) | VideoA11y (LLaVA) w/o HA&Descriptive& 5.455 & \textbf{$<$0.001} \\
 & Objective & 6.293 & \textbf{$<$0.001}	 \\
 & Accurate& 6.625 & \textbf{$<$0.001} \\
 & Clear& 5.550 & \textbf{$<$0.001} \\
\midrule
VideoA11y (GPT) | VideoA11y (LLaVA)& Descriptive& 3.953 & \textbf{$<$0.001} \\
 & Objective & 4.633 & \textbf{$<$0.001} \\
 & Accurate& 5.724 & \textbf{$<$0.001} \\
 & Clear& 4.633 & \textbf{$<$0.001} \\
\midrule
VideoA11y(GPT) | VideoA11y (GPT) w/o HA& Descriptive& 1.407		& 0.159 \\
 & Objective & 0.585 & 0.559 \\
 & Accurate& 1.312 & 0.189 \\
 & Clear& 1.455 & 0.146 \\
\bottomrule
\end{tabular}
\vspace{0.2cm} % Add some vertical space before the note
\begin{minipage}{0.82\linewidth}
\vspace{0.1cm}
\footnotesize
    Each row tests the null hypothesis that the Condition 1 and Condition 2 distributions are the same. Asymptotic significances (2-sided tests) are displayed. The significance level is 0.05.
\end{minipage}
\end{table*}

\subsection{Study 3: Statistical Analysis}
\label{app:stat_study3}
Table~\ref{tab:abaltion_pairwise_comparisons} provides additional statistical analysis for Study 3 (Section~\ref{sec:study3}). We applied a Wilcoxon Signed-Rank test to compare the performance of VideoA11y vs.\ high-quality Human Annotations for the four metrics. Significant results ($p<0.05$) are highlighted in bold. The analysis reveals that VideoA11y shows statistically significant superiority on the clear metric compared to Human Annotation. 

\subsection{Study 4: Statistical Analysis}
\label{app:stat_study4}
Table~\ref{tab:abaltion_pairwise_comparisons_experts} provides additional statistical analysis for Study 4 (Section~\ref{sec:studypd}). We applied a Wilcoxon Signed-Rank test to compare the performance of VideoA11y vs.\ high-quality Human Annotations for the four metrics. The analysis reveals that there are no statistically significant differences ($p>0.05$) in all four metrics between VideoA11y and human descriptions, likely due to the small sample size. The medium to large effect sizes for three metrics ($0.459$--$0.640$) suggest the difference in the ratings is practically important.

\section{Demographic Information of Participants}

\subsection{Demographic Information of Professional Audio Describers}
\label{app:demographic_expert}
Table~\ref{tab:participants_pro} shows the demographic information of seven professional audio describes in Study 4 (Section~\ref{sec:studypd}).

\subsection{Demographic Information of BLV Participants}
\label{app:demographic_blv}
Table~\ref{tab:participants_blv} shows the demographic information of 40 BLV participants in Study 5 (Section~\ref{sec:studyblv}). The participants include 28 males, 11 females, and 1 individual who prefers not to specify their gender. The age range spans from 18 to 51 years old. The majority are from the United States (39 participants), with one participant from the United Kingdom.

\section{Qualitative Results}
\label{app:qualitative}

\begin{figure*}[h!]
    %\vspace{-0.2cm}
    \centering
    \includegraphics[width=0.90\linewidth]{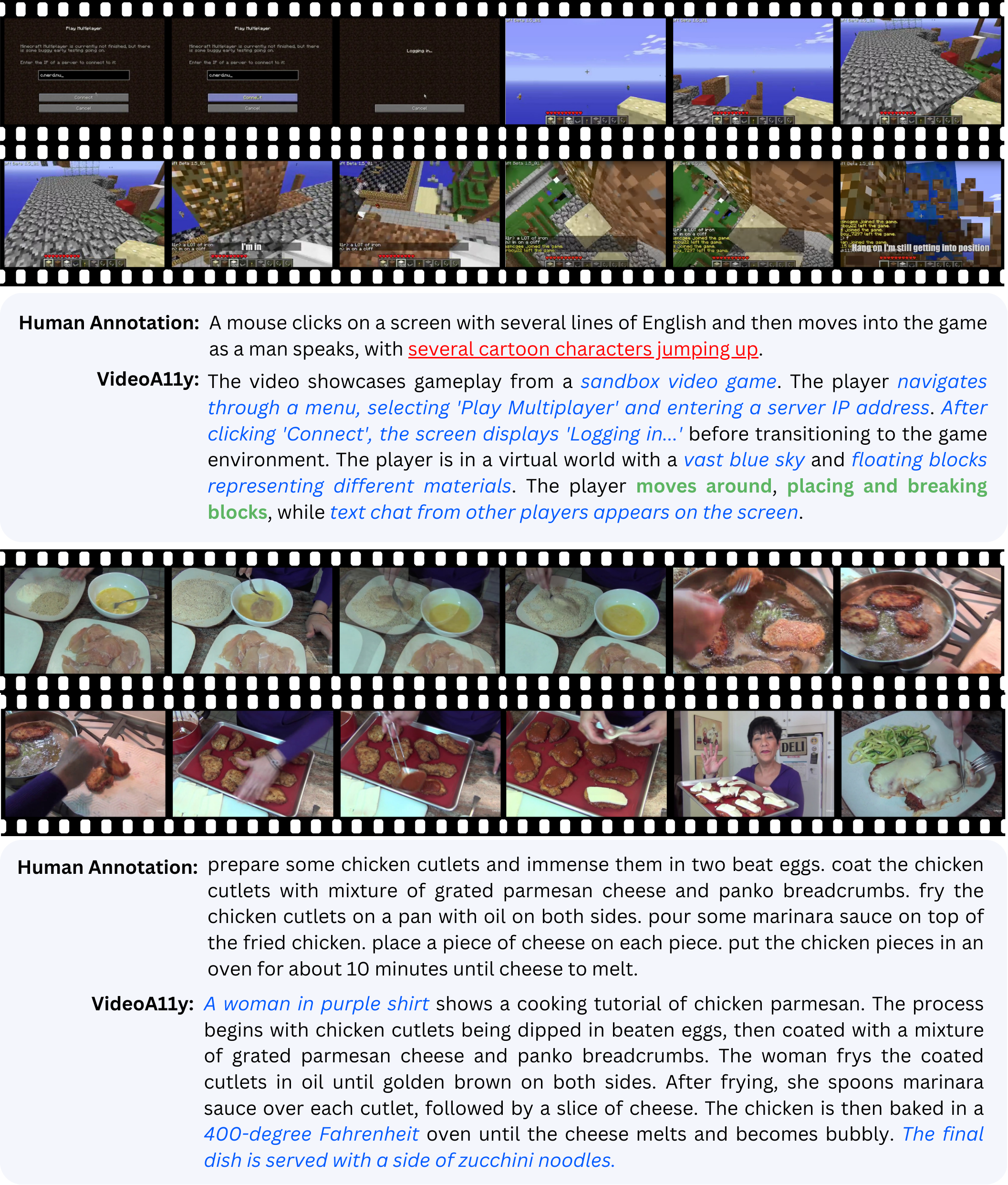}
    \caption{Qualitative comparison between human annotations and descriptions generated by VideoA11y. \textcolor{Red}{\underline{Red underline}} indicates the errors in human annotations, \textcolor{OliveGreen}{\textbf{green bold}} indicates the corrected facts, and \textit{\textcolor{Blue}{blue italics}} indicates additional details.}
    \label{fig:qualitative}
    %\Description{The figure contains two rows of screenshots depicting scenes from a video game and a cooking tutorial. Below each set of screenshots are detailed descriptions from human annotators and VideoA11y.}
    \Description{The figure contains two rows of screenshots depicting scenes from a video game and a cooking tutorial. Below each set of screenshots are detailed descriptions from human annotators and VideoA11y. In the first example, the human annotation states: "A mouse clicks on a screen with several lines of English and then moves into the game as a man speaks, with several cartoon characters jumping up", where the phrase "several cartoon characters jumping up" is underlined in red. VideoA11y's description is: "The video showcases gameplay from a sandbox video game. The player navigates through a menu, selecting 'Play Multiplayer' and entering a server IP address. After clicking 'Connect,' the screen displays 'Logging in...' before transitioning to the game environment. The player is in a virtual world with a vast blue sky and floating blocks representing different materials. The player moves around, placing and breaking blocks, while text chat from other players appears on the screen", where the phrases "sandbox video game", "navigates through a menu, selecting 'Play Multiplayer' and entering a server IP address. After clicking 'Connect,' the screen displays 'Logging in...'", "vast blue sky", "floating blocks representing different materials", and "text chat from other players appears on the screen" are italics in blue. The phrase "moves around, placing and breaking blocks" is bold in green. For the second example, the human annotation states: "Prepare some chicken cutlets and immerse them in two beat eggs. Coat the chicken cutlets with a mixture of grated parmesan cheese and panko breadcrumbs. Fry the chicken cutlets on a pan with oil on both sides. Pour some marinara sauce on top of the fried chicken. Place a piece of cheese on each piece. Put the chicken pieces in an oven for about 10 minutes until cheese melts." VideoA11y's description is: "A woman in a purple shirt shows a cooking tutorial of chicken parmesan. The process begins with chicken cutlets being dipped in beaten eggs, then coated with a mixture of grated parmesan cheese and panko breadcrumbs. The woman fries the coated cutlets in oil until golden brown on both sides. After frying, she spoons marinara sauce over each cutlet, followed by a slice of cheese. The chicken is then baked in a 400-degree Fahrenheit oven until the cheese melts and becomes bubbly. The final dish is served with a side of zucchini noodles", where the phrases "A woman in purple shirt", "400-degree Fahrenheit", and "The final dish is served with a side of zucchini noodles" are italics in blue.}
\end{figure*}

\begin{figure*}[h!]
    \centering
    \includegraphics[width=0.802\linewidth]{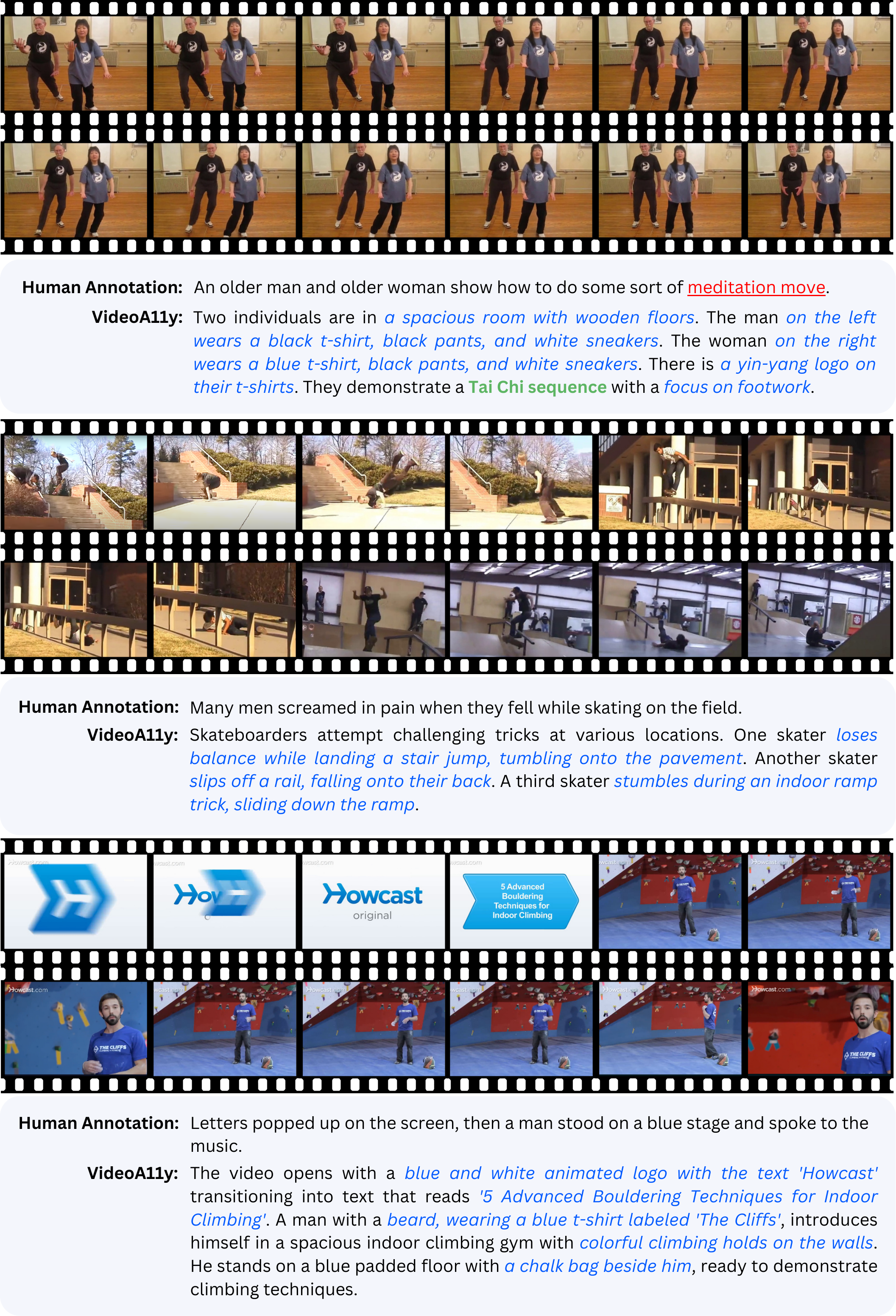}
    \vspace{-0.2cm}
    \caption{Qualitative comparison between human annotations and descriptions generated by VideoA11y. \textcolor{Red}{\underline{Red underline}} indicates the errors in human annotations, \textcolor{OliveGreen}{\textbf{green bold}} indicates the corrected facts, and \textit{\textcolor{Blue}{blue italics}} indicates additional details.}
    \label{fig:qualitative_two}
    \Description{The figure presents three video examples comparing descriptions from human annotations and VideoA11y. Below each set of screenshots are detailed descriptions from human annotators and VideoA11y. In the first video, which depicts a Tai Chi sequence, the human annotation is: "An older man and older woman show how to do some sort of meditation move", where the phrase "meditation move" is underlined in red. VideoA11y's description is: "Two individuals are in a spacious room with wooden floors. The man on the left wears a black t-shirt, black pants, and white sneakers. The woman on the right wears a blue t-shirt, black pants, and white sneakers. There is a yin-yang logo on their t-shirts. They demonstrate a Tai Chi sequence with a focus on footwork", where the phrases "a spacious room with wooden floors", "on the left wears a black t-shirt, black pants, and white sneakers", "a yin-yang logo on their t-shirts", and "focus on footwork" are italics in blue. The phrase "Tai Chi sequence" is bold in green. In the second video, showing skateboarders, the human annotation is: "Many men screamed in pain when they fell while skating on the field." VideoA11y's description is: "Skateboarders attempt challenging tricks at various locations. One skater loses balance while landing a stair jump, tumbling onto the pavement. Another skater slips off a rail, falling onto their back. A third skater stumbles during an indoor ramp trick, sliding down the ramp", where the phrases "loses balance while landing a stair jump, tumbling onto the pavement", "slips off a rail, falling onto their back", and "stumbles during an indoor ramp trick, sliding down the ramp" are italics in blue. In the third video, featuring a climbing tutorial, the human annotation is: "Letters popped up on the screen, then a man stood on a blue stage and spoke to the music." VideoA11y's description is: "The video opens with a blue and white animated logo with the text 'Howcast', transitioning into text that reads '5 Advanced Bouldering Techniques for Indoor Climbing'. A man with a beard, wearing a blue t-shirt labeled 'The Cliffs', introduces himself in a spacious indoor climbing gym with colorful climbing holds on the walls. He stands on a blue padded floor with a chalk bag beside him, ready to demonstrate climbing techniques", where the phrases "blue and white animated logo with the text 'Howcast'", "'5 Advanced Bouldering Techniques for Indoor Climbing'", "beard, wearing a blue t-shirt labeled 'The Cliffs'", "colorful climbing holds on the walls", and "a chalk bag beside him" are italics blue.}
\end{figure*}

Figure~\ref{fig:qualitative} and ~\ref{fig:qualitative_two} illustrate qualitative examples where we compare Human Annotations with the descriptions generated by VideoA11y. 

\begin{figure*}[h!]
    \centering
     \includegraphics[width=0.90\linewidth]{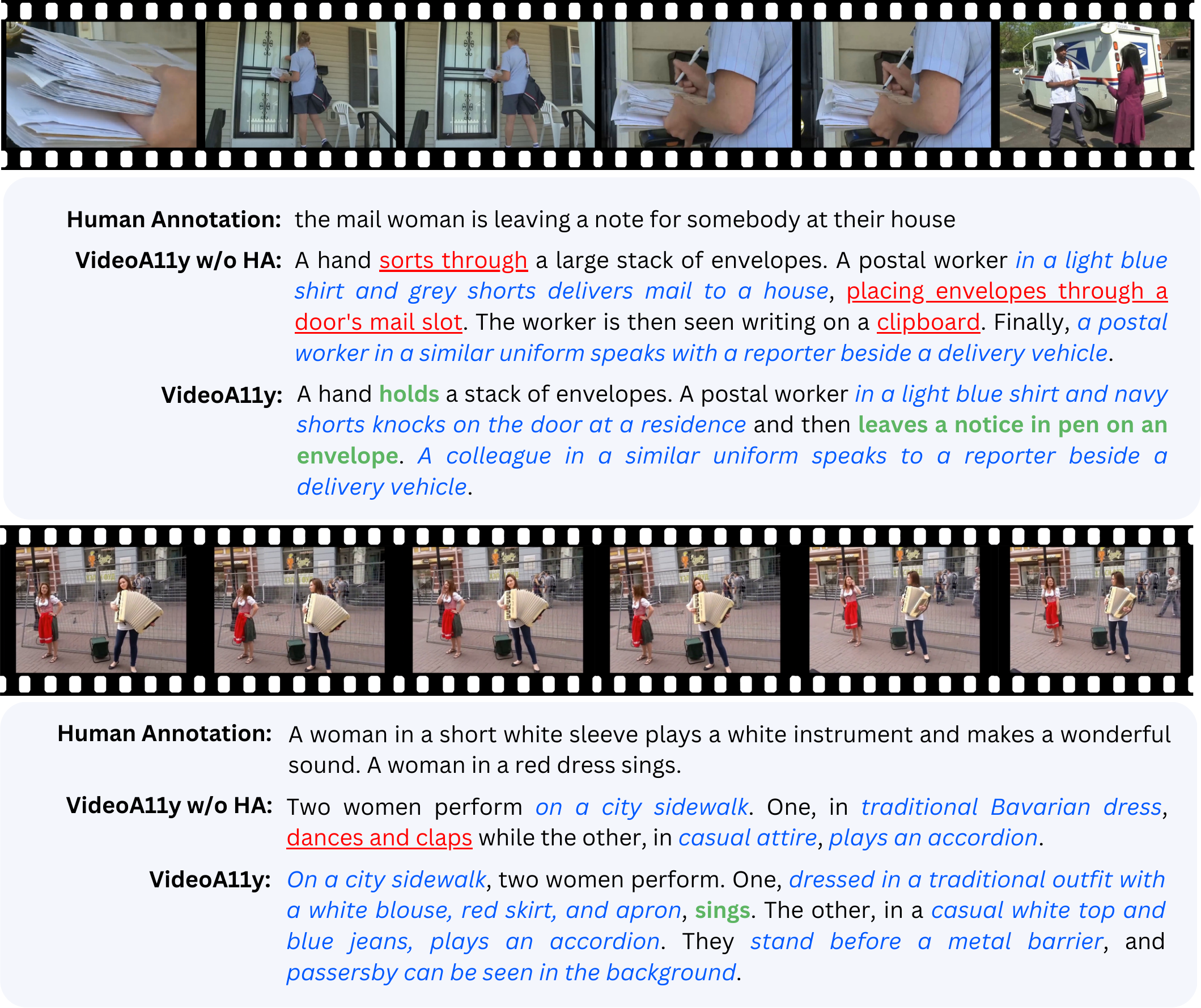}
    \caption{Qualitative comparison between human annotations, VideoA11y w/o HA, and VideoA11y. HA: Human Annotation. \textcolor{Red}{\underline{Red underline}} indicates the hallucinations from VideoA11y w/o HA, \textcolor{OliveGreen}{\textbf{green bold}} indicates the corrected facts, and \textit{\textcolor{Blue}{blue italics}} denotes correct additional details that are absent in the human annotations.}
    \label{fig:hallucination}
    \Description{The figure illustrates comparative analysis of descriptions from Human Annotations, VideoA11y w/o HA, and VideoA11y. In the first example, showing a postal worker, the human annotation is: "the mail woman is leaving a note for somebody at their house." The description from VideoA11y w/o HA is: "A hand sorts through a large stack of envelopes. A postal worker in a light blue shirt and grey shorts delivers mail to a house, placing envelopes through a door's mail slot. The worker is then seen writing on a clipboard. Finally, a postal worker in a similar uniform speaks with a reporter beside a delivery vehicle", where the phrases "sorts through", "placing envelopes through a door's mail slot", and "clipboard" are underlined in red. The phrases "in a light blue shirt and grey shorts delivers mail to a house", and "a postal worker in a similar uniform speaks with a reporter beside a delivery vehicle" are italics in blue. VideoA11y's description is: "A hand holds a stack of envelopes. A postal worker in a light blue shirt and navy shorts knocks on the door at a residence and then leaves a notice in pen on an envelope. A colleague in a similar uniform speaks to a reporter beside a delivery vehicle", where the phrases "holds", and "leaves a notice in pen on an envelope" are bold in green. The phrases "in a light blue shirt and navy shorts knocks on the door at a residence", and "A colleague in a similar uniform speaks to a reporter beside a delivery vehicle" are italics in blue. In the second example, which shows two women performing on a sidewalk, the human annotation is: "A woman in a short white sleeve plays a white instrument and makes a wonderful sound. A woman in a red dress sings." The description from VideoA11y w/o HA is: "Two women perform on a city sidewalk. One, in traditional Bavarian dress, dances and claps while the other, in casual attire, plays an accordion", where the phrase "dances and claps" is underlined in red. The phrases "on a city sidewalk", "traditional Bavarian dress", and "casual attire, plays an accordion" are italics in blue. VideoA11y's description is: "On a city sidewalk, two women perform. One, dressed in a traditional outfit with a white blouse, red skirt, and apron, sings. The other, in a casual white top and blue jeans, plays an accordion. They stand before a metal barrier, and passersby can be seen in the background", where the phrase "sings" is bold green. The phrase "On a city sidewalk", "dressed in a traditional outfit with a white blouse, red skirt, and apron", "casual white top and blue jeans, plays an accordion", "stand before a metal barrier", and "passersby can be seen in the background" are italics in blue.}
\end{figure*}

Figure~\ref{fig:hallucination} illustrates examples of hallucination phenomena observed in the descriptions generated by VideoA11y w/o HA. This figure provides a comparative analysis of descriptions from Human Annotators, VideoA11y w/o HA, and VideoA11y. In the absence of human annotation as a reference, GPT-4V sometimes introduces actions or details that are not present in the video or provide incorrect information. For instance, for the first video, the model described actions such as ``sorts through'' and ``placing envelopes through a door's mail slot'', which are not in the video content. Furthermore, for the last video, the movements of Tai Chi were incorrectly classified as a ``dance routine'' without the hint from human annotations.

\begin{figure*}[h!]
    \centering
     \includegraphics[width=0.835\linewidth]{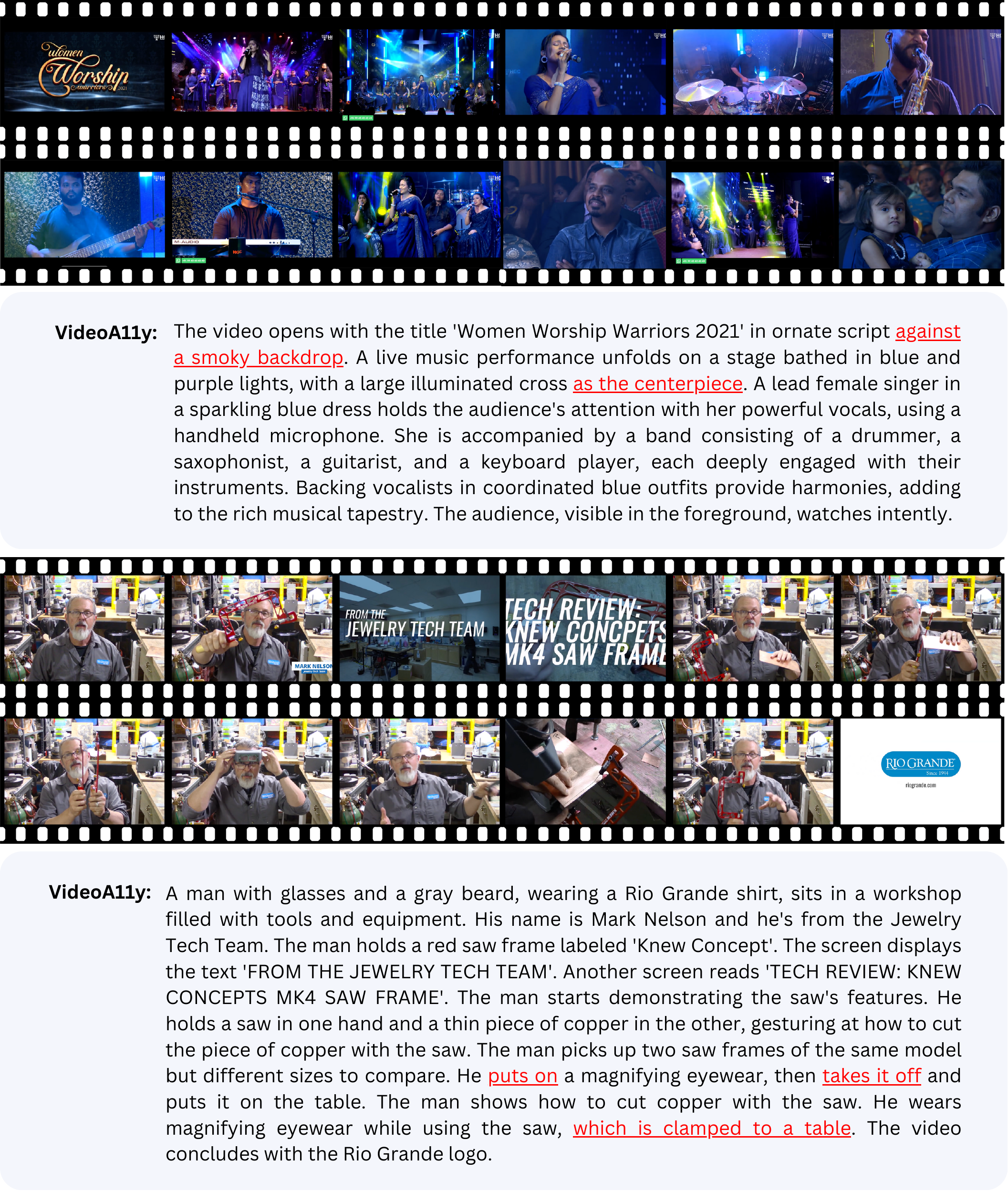}
    \caption{Examples of minor inaccuracies in descriptions generated by VideoA11y \textcolor{Red}{\underline{Red underline}} indicates the inaccuracies content in the descriptions.}
    \label{fig:inaccuracies}
    \Description{The figure minor inaccuracies in descriptions generated by VideoA11y. In the first example, the description is: "The video opens with the title 'Women Worship Warriors 2021' in ornate script against a smoky backdrop. A live music performance unfolds on a stage bathed in blue and purple lights, with a large illuminated cross as the centerpiece. A lead female singer in a sparkling blue dress holds the audience's attention with her powerful vocals, using a handheld microphone. She is accompanied by a band consisting of a drummer, a saxophonist, a guitarist, and a keyboard player, each deeply engaged with their instruments. Backing vocalists in coordinated blue outfits provide harmonies, adding to the rich musical tapestry. The audience, visible in the foreground, watches intently.", "against a smoky backdrop", and "as the centerpiece" are underlined in red. In the second example, the description is: "A man with glasses and a gray beard, wearing a Rio Grande shirt, sits in a workshop filled with tools and equipment. His name is Mark Nelson and he's from the Jewelry Tech Team. The man holds a red saw frame labeled 'Knew Concept'. The screen displays the text 'FROM THE JEWELRY TECH TEAM'. Another screen reads 'TECH REVIEW: KNEW CONCEPTS MK4 SAW FRAME'. The man starts demonstrating the saw's features. He holds a saw in one hand and a thin piece of copper in the other, gesturing at how to cut the piece of copper with the saw. The man picks up two saw frames of the same model but different sizes to compare. He puts on a magnifying eyewear, then takes it off and puts it on the table. The man shows how to cut copper with the saw. He wears magnifying eyewear while using the saw, which is clamped to a table. The video concludes with the Rio Grande logo.", "puts on", "takes it off", and "which is clamped to a table" are underlined in red.}
\end{figure*}

Figure~\ref{fig:inaccuracies} illustrates examples of minor inaccuracies in descriptions generated by VideoA11y. For the first example, the phrase ``against a smoky backdrop" is a hallucination, as no smoky backdrop is present in the actual video. Additionally, the description ``a large illuminated cross as the centerpiece" is somewhat misleading, as the cross is part of the stage's backdrop rather than being the central focus of the scene. In the second example, the man does not ``put on" and ``take off" magnifying eyewear but merely gestures with it. Furthermore, the statement ``He wears magnifying eyewear while using the saw, which is clamped to a table" is incorrect, as the copper sheet—not the saw—is clamped to the table.

\clearpage

\begin{table*}[t]
\footnotesize
\centering
\caption{Overall pairwise comparisons between VideoA11y and other methods in Study 2. HA: Human Annotation.}
\label{tab:pairwise_comparisons_2}
\begin{tabular}{>{\raggedright\arraybackslash}p{0.35\linewidth}
>{\raggedright\arraybackslash}p{0.14\linewidth}
>{\raggedright\arraybackslash}p{0.11\linewidth}
>{\raggedright\arraybackslash}p{0.08\linewidth}}
\toprule
\textbf{Condition 1 | Condition 2} & \textbf{Metric} & \textbf{Test Statistics} & \textbf{P Value} \\
\midrule
GPT-4V | Human Annotation& Descriptive& 3.033  & \textbf{0.002} \\
 & Objective & 1.129 & 0.259 \\
 & Accurate& 0.738 & 0.461 \\
 & Clear& 1.381 & 0.167 \\
\midrule
GPT-4V w/ HA | Human Annotation& Descriptive& 1.760  & 0.078 \\
 & Objective & 1.313 & 0.189 \\
 & Accurate& 0.962 & 0.336 \\
 & Clear& 0.079 & 0.937 \\
\midrule
% GPT-4V w/ HA | GPT-4V& Descriptive& 1.274  & 0.203 \\
%  & Objective & 0.184 & 0.854 \\
%  & Accurate& 0.224 & 0.823 \\
%  & Clear& 1.302 & 0.193 \\
% \midrule
VideoA11y w/o HA | Human Annotation& Descriptive& 5.515  & \textbf{$<$0.001} \\
 & Objective & 4.228 & \textbf{$<$0.001} \\
 & Accurate& 4.401 & \textbf{$<$0.001} \\
 & Clear& 5.090 & \textbf{$<$0.001} \\
\midrule
VideoA11y w/o HA | GPT-4V& Descriptive& 2.482 & \textbf{0.013} \\
 & Objective & 3.099 & \textbf{0.002} \\
 & Accurate& 3.663 & \textbf{$<$0.001} \\
 & Clear& 3.709 & \textbf{$<$0.001} \\
\midrule
VideoA11y w/o HA | GPT-4V w/ HA& Descriptive& 3.755 & \textbf{$<$0.001} \\
 & Objective& 2.915 & \textbf{0.004} \\
 & Accurate& 3.439 & \textbf{$<$0.001} \\
 & Clear& 5.011 & \textbf{$<$0.001} \\
\midrule
VideoA11y | Human Annotation& Descriptive& 7.156 & \textbf{$<$0.001} \\
 & Objective & 6.066 & \textbf{$<$0.001}	 \\
 & Accurate& 6.483 & \textbf{$<$0.001} \\
 & Clear& 8.116 & \textbf{$<$0.001} \\
\midrule
VideoA11y | GPT-4V& Descriptive& 4.123 & \textbf{$<$0.001} \\
 & Objective & 4.937 & \textbf{$<$0.001} \\
 & Accurate& 5.745 & \textbf{$<$0.001} \\
 & Clear& 6.735 & \textbf{$<$0.001} \\
\midrule
VideoA11y | GPT-4V w/ HA& Descriptive& 5.397 & \textbf{$<$0.001} \\
 & Objective & 4.753 & \textbf{$<$0.001}	 \\
 & Accurate& 5.521 & \textbf{$<$0.001} \\
 & Clear& 8.037 & \textbf{$<$0.001} \\
\bottomrule
\end{tabular}
\vspace{0.2cm} % Add some vertical space before the note
\begin{minipage}{0.80\linewidth}
\vspace{0.1cm}
\footnotesize
    Each row tests the null hypothesis that the Condition 1 and Condition 2 distributions are the same. Asymptotic significances (2-sided tests) are displayed. The significance level is 0.05.
\end{minipage}
\end{table*}

\begin{table*}[ht]
\footnotesize
\centering
\caption{Overall pairwise comparisons between VideoA11y and trained human descriptions in Study 3.}
\label{tab:abaltion_pairwise_comparisons}
%\vspace{-2mm}
\begin{tabular}{>{\raggedright\arraybackslash}p{0.30\linewidth}
>{\raggedright\arraybackslash}p{0.10\linewidth}
>{\raggedright\arraybackslash}p{0.1\linewidth}
>
{\raggedright\arraybackslash}p{0.08\linewidth}}
\toprule
\textbf{Condition 1 | Condition 2} & \textbf{Metric} & \textbf{Test Statistics} & \textbf{P Value}  \\
\midrule
VideoA11y | Human Annotation& Descriptive & 0.551  & 0.582 \\
\midrule
VideoA11y | Human Annotation& Objective & 0.238 & 0.812 \\
\midrule
VideoA11y | Human Annotation& Accurate & 1.191 & 0.234 \\
\midrule
VideoA11y | Human Annotation& Clear & 2.843 & \textbf{0.004} \\
\bottomrule
\end{tabular}
\begin{minipage}{0.8\linewidth}
\vspace{0.1cm}
\footnotesize
    Each row tests the null hypothesis that the Condition 1 and Condition 2 distributions are the same. Asymptotic significances (2-sided tests) are displayed. The significance level is 0.05.
\end{minipage}
\end{table*}

\begin{table*}[h!]
\footnotesize
\centering
\caption{Overall pairwise comparisons between VideoA11y and trained human descriptions in Study 4.}
\label{tab:abaltion_pairwise_comparisons_experts}
%\vspace{-2mm}
\begin{tabular}{>{\raggedright\arraybackslash}p{0.30\linewidth}
>{\raggedright\arraybackslash}p{0.10\linewidth}
>{\raggedright\arraybackslash}p{0.1\linewidth}
>{\raggedright\arraybackslash}p{0.1\linewidth}
>
{\raggedright\arraybackslash}p{0.08\linewidth}}
\toprule
\textbf{Condition 1 | Condition 2} & \textbf{Metric} & \textbf{Effect Size} & \textbf{Test Statistics} & \textbf{P Value}  \\
\midrule
VideoA11y | Human Annotation& Descriptive & 0.459 & 1.214  & 0.225 \\
\midrule
VideoA11y | Human Annotation& Objective & 0.288 & 0.762 & 0.446 \\
\midrule
VideoA11y | Human Annotation& Accurate & 0.515 & 1.363 & 0.173 \\
\midrule
VideoA11y | Human Annotation& Clear & 0.640 & 1.693 & 0.090 \\
\bottomrule
\end{tabular}
\begin{minipage}{0.8\linewidth}
\vspace{0.1cm}
\footnotesize
    Each row tests the null hypothesis that the Condition 1 and Condition 2 distributions are the same. Asymptotic significances (2-sided tests) are displayed. The significance level is 0.05.
\end{minipage}
\end{table*}\begin{table*}[b!]
\centering
\footnotesize
\caption{Participant demographics of professional audio describes in Study 4.}
\label{tab:participants_pro}
\begin{tabular}{@{}cllllll@{}}
\toprule
\textbf{Pseudonym} & \textbf{Age} & \textbf{Gender} & \textbf{Ethnicity} & \textbf{Race} & \textbf{Country} & \textbf{Years of Experience} \\ 
\midrule
P1  & 50 & Female & Not Hispanic or Latino & White & United States & 7\\
P2  & 32 & Female & Not Hispanic or Latino & Black & United States & 5\\
P3  & 33 & Female & Not Hispanic or Latino & White & United States & 4\\
P4  & 30 & Male & Not Hispanic or Latino & White & Canada & 3\\
P5  & 26 & Non-Binary & Not Hispanic or Latino & White & United States & 4\\
P6  & 66 & Female & Not Hispanic or Latino & White & United States & 23\\
P7  & 30 & Male & Not Hispanic or Latino & Black & United States & 5\\
\bottomrule
\end{tabular}
\end{table*}

\begin{table*}[t]
\centering
\footnotesize
\caption{Participant demographics in the BLV study. The description of vision is self-reported by participants.}
\label{tab:participants_blv}
\begin{tabular}{@{}cllllll@{}}
\toprule
\textbf{Pseudonym} & \textbf{Age} & \textbf{Gender} & \textbf{Ethnicity} & \textbf{Race} & \textbf{Country} & \textbf{BLV Level} \\ 
\midrule
P1  & 27 & Male & Not Hispanic or Latino & Black & United States & Legally Blind (20/800) \\
P2  & 28 & Male & Not Hispanic or Latino & Black & United States & Legally Blind (20/1000) \\
P3  & 28 & Male & Not Hispanic or Latino & White & United States & Legally Blind (20/500 - 20/1000) \\
P4  & 26 & Male & Not Hispanic or Latino & Black & United States & Legally Blind (20/700) \\
P5  & 28 & Male & Not Hispanic or Latino & Black & United States & Legally Blind (20/500) \\
P6  & 27 & Male & Not Hispanic or Latino & White & United States & Legally Blind (20/500) \\
P7  & 44 & Male & Not Hispanic or Latino & Asian & United States & Totally Blind \\
P8  & 33 & Unknown & Hispanic or Latino & More than one race & United States & Totally Blind \\
P9  & 32 & Female & Hispanic or Latino & Unknown & United States & Totally blind \\
P10 & 23 & Female & Hispanic or Latino & Black & United States & Legally Blind (20/1000) \\
P11 & 26 & Male & Not Hispanic or Latino & Black & United States & Legally Blind (20/500) \\
P12 & 24 & Male & Not Hispanic or Latino & Black & United States & Legally Blind (20/500) \\
P13 & 29 & Male & Not Hispanic or Latino & White & United States & Legally Blind (20/500 - 20/1000) \\
P14 & 26 & Male & Not Hispanic or Latino & Black & United States & Legally Blind (20/900) \\
P15 & 22 & Female & Hispanic or Latino & Unknown & United States & Totally blind \\
P16 & 31 & Male & Not Hispanic or Latino & White & United States & Legally Blind (20/700) \\
P17 & 26 & Male & Not Hispanic or Latino & Black & United States & Totally Blind \\ 
P18 & 23 & Female & Not Hispanic or Latino & White & United States & Legally Blind (20/200) \\
P19 & 20 & Male & Not Hispanic or Latino & Unknown & United States & Legally Blind (20/400) \\
P20 & 21 & Female & Hispanic or Latino & Black & United States & Legally Blind (20/1000) \\
P21 & 28 & Female & Not Hispanic or Latino & Black & United States & Legally Blind (20/600) \\
P22 & 25 & Male & Not Hispanic or Latino & Black & United States & Legally Blind (20/200) \\
P23 & 23 & Male & Not Hispanic or Latino & White & United States & Legally Blind (20/200) \\
P24 & 27 & Female & Not Hispanic or Latino & White & United States & Legally Blind (20/200) \\
P25 & 28 & Male & Not Hispanic or Latino & White & United States & Legally Blind (20/200) \\
P26 & 21 & Male & Hispanic or Latino & White & United States & Legally Blind (20/500) \\
P27 & 24 & Male & Not Hispanic or Latino & White & United States & Legally Blind (20/600) \\
P28 & 20 & Female & Hispanic or Latino & White & United States & Legally Blind (20/500) \\
P29 & 21 & Male & Not Hispanic or Latino & Black & United States & Legally Blind (20/400) \\
P30 & 21 & Female & Not Hispanic or Latino & Black & United States & Legally Blind (20/500) \\
P31 & 22 & Male & Not Hispanic or Latino & Black & United States & Legally Blind (20/400) \\
P32 & 21 & Male & Hispanic or Latino & White & United States & Legally Blind (20/500) \\
P33 & 20 & Male & Not Hispanic or Latino & Black & United States & Legally Blind (20/500) \\
P34 & Unknown & Male & Not Hispanic or Latino & Black & United States & Legally Blind (20/400) \\
P35 & 30 & Male & Hispanic or Latino & White & United States & Legally Blind (20/400) \\
P36 & 27 & Female & Not Hispanic or Latino & Black & United States & Legally Blind (20/400) \\
P37 & 31 & Male & Not Hispanic or Latino & Black & United States & Legally Blind (20/500) \\
P38 & 27 & Male & Not Hispanic or Latino & White & United Kingdom & Legally Blind (20/1000) \\
P39 & 18 & Male & Not Hispanic or Latino & Black & United States & Legally Blind (20/500) \\
P40 & 51 & Female & Not Hispanic or Latino & White & United States & Totally Blind \\
\bottomrule
\end{tabular}
\end{table*}

\clearpage

\clearpage

%TC:endignore
\end{document}